\documentclass[11pt]{birkjour_t2}

\usepackage{amsmath,amssymb,amsfonts}
\usepackage{multirow}
\usepackage{tikz}
\usepackage{listings}
\usepackage{subfigure}
\usepackage{curves}
\usepackage{xypic}  
\usepackage[mathscr]{eucal}
\usepackage[pdftex]{hyperref}
\usepackage[all]{xy}
\usepackage{graphics,graphicx}
\usepackage{epsfig}
\usepackage{color}
\definecolor{red-}{rgb}{1.0,0.0,0.0}
\definecolor{grey}{rgb}{0.6, 0.6, 0.6}
\definecolor{magenta}{rgb}{1, 0, 1}
\definecolor{brown}{rgb}{0.5,0.2,0.0}
\definecolor{brown-}{rgb}{0.0,0.1,1.0}
\definecolor{green-}{rgb}{0.0, 0.6, 0.0}
\definecolor{gold}{rgb}{0.8,0.7,0.0}
\definecolor{black}{rgb}{0.0,0.0,0.0}
\definecolor{DarkGreen}{rgb}{0.0,0.3,0.2}
\definecolor{LightGreen}{rgb}{0.8,1.0, 0.8}
\definecolor{yellow}{rgb}{0.9,0.9,0.0}
\definecolor{blue-}{rgb}{0.0,0.1,1.0}

\usepackage{bm}  
\usepackage{cancel}
\newcommand\Area{\rule[-4mm]{0mm}{10mm}}
\newcommand\quadrato{{\hspace*{\fill}$\Box$\vskip12pt plus 1pt}}

\newcommand\gra{\alpha}

\newcommand\reals{{\mathbb R}}
\newcommand\comp{{\mathbb C}}

\newcommand\pn[1]{{\mathbb P}^{#1}}

\newcommand{\x}{{\bm x}}

\newcommand\R{{\mathbb R}}
\newcommand\C{{\mathbb C}}

\newcommand{\be}{\begin{equation}}
\newcommand{\ee}{\end{equation}}

\newtheorem{theorem}{Theorem}[section]
\newtheorem{lemma}[theorem]{Lemma}

\newtheorem{prop}[theorem]{Proposition}

\newtheorem{definition}[theorem]{Definition}
\newtheorem{re}[theorem]{Remark}
\newtheorem{defre}[theorem]{Definition--Remark}
\newtheorem{pargrph}[theorem]{}
\newtheorem{examp}[theorem]{Example}

\newtheorem{propdef}[theorem]{Proposition--Definition}

\def\cocoa{{\hbox{\rm C\kern-.13em o\kern-.07em C\kern-.13em o\kern-.15em A}}}

\newenvironment{rem*}{\begin{re}\em}{\end{re}}
\newenvironment{example*}{\begin{examp}\em}{\end{examp}}
\newenvironment{definition*}{\begin{definition}\em}{\end{definition}}

\newenvironment{prgrph*}[1]{\indent\begin{pargrph}{\bf #1.}\em\
}{\end{pargrph}}
\newenvironment{defre*}{\begin{defre}\em}{\end{defre}}
\newenvironment{MM*}{\begin{MM}\em}{\end{MM}}

\begin{document}

\title{Geometry of the Hough transforms  with  applications to  synthetic data\footnote{
\newline 
\noindent M.C. Beltrametti,
Dipartimento di Matematica,
Universit\`a degli Studi di Genova, Genova, Italy. 
e-mail {\tt beltrame@dima.unige.it}
 \newline
\noindent C. Campi,
Dipartimento di Medicina DIMED,
Universit\`a degli Studi di Padova, Padova Italy. 
e-mail {\tt cristina.campi@unipd.it}
 \newline
\noindent A.M. Massone,
Dipartimento di Matematica,
Universit\`a degli Studi di Genova, Genova, Italy. 
e-mail {\tt massone@dima.unige.it}
 \newline
\noindent
 M. Torrente,
Dipartimento di Economia,
Universit\`a degli Studi di Genova, Genova, Italy. 
e-mail {\tt marialaura.torrente@economia.unige.it}
}}

\author{M.C. Beltrametti, C. Campi, A.M. Massone, and  M. Torrente}

\maketitle

\begin{abstract} In the framework of the Hough transform technique to detect  curves in images, we provide a bound for the number of Hough transforms  to be considered  for a successful optimization of the accumulator function  in the recognition algorithm. Such a  bound is consequence of  geometrical arguments.
We  also show the robustness of the  results when applied to synthetic  datasets strongly perturbed by noise. 
An algebraic approach, discussed in  the appendix, leads to a better bound of theoretical interest in the exact case.
\end{abstract}

%

\section*{Introduction}
The Hough transform is a standard technique for feature extraction used in image analysis and digital image processing. 
Such a technique was first used to detect straight lines in images \cite{Hough}. It is based on the \emph{point-line duality} as follows: points on a straight line, defined by an equation in the image plane $\langle x, y \rangle$ with the usual natural parametrization, correspond to lines in the parameter space $\langle A, B \rangle$ that intersect in a single point. This point uniquely identifies the coefficients in the equation of the original straight line  (analogous procedures to detect circles and ellipses in images have been  introduced in~\cite{DH}). From a computational point of view, this result gives us a procedure  to recognize straight lines in $2$-dimensional images in which discontinuity regions in the image are highlighted through an {\it edge detection} algorithm; the parameter space is discretized in cells and an accumulator function is defined on it, whose maximum provides us  with the parameters' values that identify the line.

Thanks to algebraic geometrical arguments, the Hough transform definition has been extended to include special classes of curves \cite{BMP, maetal}. In \cite{BMP},  a characterization of families of irreducible algebraic plane curves of the same degree for which is defined a Hough-type correspondence is provided. In fact, given a family  $\mathcal{F}$ of algebraic curves, a general point $p$ in the image plane corresponds to an algebraic  locus, $\Gamma_p(\mathcal{F})$, in the parameter space. The families $\mathcal{F}$ such that, as $p$ varies on a given curve $\mathcal{C}$ from $\mathcal{F}$, satisfy the  {\em regularity condition} that the hypersurfaces $\Gamma_p(\mathcal{F})$ meet in one and only one point (which in turn defines the curve $\mathcal{C}$),  are called {\em Hough regular}. This paper is a sequel of \cite{BMP}. Indeed,  in \cite{BMP} the Hough transform technique was performed for the automated recognition of cubic and quartic curves, and the accuracy of the detection was tested against synthetic data. Here, the aim is to reduce the amount of the dataset to be taken into account. Furthermore, the power of this procedure is then tested on $2$-dimensional astronomical and medical real images in \cite{maetal}.


Let ${\mathcal{F}}=\{{\mathcal{C}}_{\bm\lambda}\}$  be  a Hough regular  family of algebraic plane curves ${\mathcal{C}}_{\bm\lambda}$.  Let's  highlight here the steps of the 
standard recognition algorithm, of a given profile of interest ${\mathcal P}$ in a real image,  on which the Hough transform technique for  such families of plane curves is founded.  We  refer   to \cite[Section 6]{BMP}, \cite[Section 4]{maetal}, and also \cite[Section 5]{RBM} for complete details. 

A pre-processing step of the algorithm consists of the application of a standard edge detection technique on the image
(see  \cite{CANNY}  for a detailed description of this operator). This step  reduces the number of points of interest highlighting the profile ${\mathcal{P}}$ of which one has to compute the Hough transform.
Then a discretization of the parameter space is required, which possibly exploits bounds on the parameter values computed by using either the Cartesian or the parametric form of the curve in the image space. A last step constructs the accumulator function after a discretization of the   parameter space. The value of the accumulator in a cell of the discretized space corresponds to the number of times the Hough transforms of  selected points of interest reach that cell. As a final outcome of the algorithm, the parameter values characterizing the curve best approximating the profile ${\mathcal{P}}$ in the image space  correspond to the parameter values identifying the cell where the accumulator function reaches its maximum.

Thus, in practice, the computational burden associated to the accumulator function computation and optimization leads to the need of reducing as most as possible the number of points of interest to be considered.
By using classical geometrical arguments we provide in Proposition \ref{A&Cprop}, and in an exact context,  a bound  which quite  significantly  decreases the  number  of Hough transforms  $\Gamma_{p_j}({\mathcal{F}})$ of points $p_j\in {\mathcal C}_{\bm\lambda}$ making true the regularity condition $\cap_{p_j} \Gamma_{p_j}({\mathcal{F}}) =\{\bm\lambda\}$. This suggests to significantly  bound the number of Hough transforms  to be considered to recognize curves in images. Indeed, such a  bound  applies quite efficiently  in  concrete examples,  and it turns out to be quite robust both  in presence of noisy background and against  random perturbation of points' locations, as shown in Section \ref{App1}. This significantly enhances the results of \cite[Section 6]{BMP}, with special regard  to robustness in presence of noisy background.

A better understanding  of the behavior of the equations  defining the  Hough transforms in the parameter space leads to   a refinement of Proposition \ref{A&Cprop}. This algebraic approach provides a  much better bound, called $\nu_{\rm best}$ (see Proposition \ref{finally}, Appendix A),  which appears of  purely  theoretical interest since   such a bound can  be even too strong for practical purposes. Indeed, let ${\mathcal C}_{\bm\lambda}$ be a curve from a family ${\mathcal F}$ potentially approximating  a profile 
${\mathcal P}$. Since $\nu_{\rm best}$ can be very small (for instance, $\nu_{\rm best}=2$ in the examples provided in Appendix A), random perturbations of $\nu_{\rm best}$ point's locations on ${\mathcal C}_{\bm\lambda}$  may produce a dataset of points not  properly highlighting the profile ${\mathcal P}$. 

The paper is organized as follows. In Section \ref{BP} we recall some  background  material.  Section \ref{FHT} is devoted to the proof of the bound mentioned above. We then provide several examples in Section~\ref{EX}. In Section \ref{App1}, applications  to synthetic  data for four families of curves (the same considered in \cite{BMP}) show  the efficiency and the robustness of the result. Finally, our conclusions are offered in Section \ref{Conc}.

\section{Preliminaries }\label{BP}

Most of the results in this section  hold over an infinite integral ring $K$. 
However, unless otherwise specified,  we restrict to the case of interest in the applications, assuming either  $K= \R$ or $K=\C$ the fields of real or complex numbers.

For every $t$-tuples of independent  parameters ${\bm\lambda}:=(\lambda_1,\ldots,\lambda_t)\in K^t$, let 
\begin{equation}\label{GE} 
f_{{\bm \lambda}}({\x})=\sum_{i_1,\ldots,i_n} x_1^{i_1} \ldots  x_n^{i_n}g_{i_1\ldots i_n}({\bm\lambda}) , \;\;\;\; 0\leq  i_1+\cdots+i_n\leq d,
\end{equation}
be a family ${\mathcal P}$ of  non-constant irreducible polynomials in the indeterminates ${\x}:=(x_1,\ldots,x_n)$,  
of a given degree $d$ (not depending on ${\bm\lambda}$), whose coefficients $g_{i_1\ldots i_n}({\bm\lambda})$ 
are the evaluation in ${\bm\lambda}$ of polynomials $g_{i_1\ldots i_n}({\bm\Lambda})\in K[{\bm\Lambda}]$ in a new series of indeterminates  ${\bm\Lambda}=(\Lambda_1,\ldots,\Lambda_t)$. Let
${\mathcal{H}}_{\bm\lambda} = \{\x \in {\mathbb A}_{\x}^n(K) \:|\: f_{\bm\lambda}(\x)=0\} $, 
and let assume that ${\mathcal{H}}_{\bm\lambda}$ is a hypersurface for each parameter ${\bm\lambda}$ belonging to a Euclidean open subset ${\mathcal{U}}\subseteq K^t$
(of course, this is always  the  case if the base field  is $K=\C$). 
Clearly, if $K=\C$, such hypersurfaces are irreducible, that is, they consist of a single component, 
 since  the polynomials of the family ${\mathcal P}$ 
are assumed to be irreducible in $K[{\x}]$. If $K=\R$, the case of interest in the applications, we assume that ${\mathcal H}_{\bm\lambda}$ is a{ \em real hypersurface}, that is, a hypersurface over $\C$ with a real $(n-1)$-dimensional component  in the affine space  ${\mathbb A}_{\x}^n(\R)$  (see \cite[Theorem 4.5.1]{BCR} for  explicit conditions equivalent to our assumption). 

Since the polynomials
$f_{\bm\lambda}(\x)$ are irreducible over $K$, the zero loci ${\mathcal{H}}_{\bm\lambda}$ are irreducible up to components of dimension $\leq n-2$, that is, they consist of  a single $(n-1)$-dimensional component (see \cite[Remark 1.5]{RBM}  for related comments in the cases $n=2$ and $n=3$, respectively).

So, we assume $\mathcal{F}$ to be a  
{\em family of irreducible hypersurfaces $($with possibly a finite set of lower dimensional components$)$ which share the degree}.

\begin{definition*}\label{def1}
Let  $\mathcal{F}$ be a family of hypersurfaces  ${\mathcal{H}}_{\bm\lambda}$ as above, 
and let $p=(x_1(p),\ldots, x_n(p))$ be a point in the {\em image space}  ${\mathbb A}_{\x}^n(K)$. 
Let  ${\Gamma}_p(\mathcal{F})$ be the locus defined in the affine $t$-dimensional {\em parameter space} 
${\mathbb A}_{\bm \Lambda}^t(K)$ by the polynomial equation 
$$f_p(\bm \Lambda) =
\sum_{i_1,\ldots,i_n} x_1(p)^{i_1} \ldots  x_n(p)^{i_n}g_{i_1\ldots i_n}({\bm \Lambda})=0, \;\;\;\;\;\; 0\leq i_1+\cdots+i_n\leq d. 
$$ 
We say that ${\Gamma}_p(\mathcal{F})$ is the {\em Hough transform of the point $p$ 
with respect to the family} $\mathcal{F}$.  
If no confusion will arise, we simply say that ${\Gamma}_p(\mathcal{F})$ 
is the {\em Hough transform of $p$}. 
\end{definition*}

See also Appendix \ref{AB} for more details on degree and dimension of the Hough transform.

Summarizing,  the polynomials family defined by (\ref{GE})  leads to a polynomial 
$F({\x}; {\bm \Lambda})\in K[{\x}; {\bm \Lambda}]$  whose evaluations at  points ${\bm\lambda} \in{\mathcal{U}}$ 
and  $p=(x_1(p), \ldots, x_n(p)) \in {\mathbb A}_{\x}^n(K)$ give back the equations of ${\mathcal{H}}_{\bm\lambda}$ and ${\Gamma}_p(\mathcal{F})$, respectively. That is,  
$$
{\mathcal{H}}_{\bm\lambda} : F({\x};{\bm\lambda})=f_{\bm\lambda}({\x})=0 \;\;\;\;
{\rm and}
\;\;\;\;
{\Gamma}_p(\mathcal{F}) : F(p; {\bm \Lambda})=f_p(\bm \Lambda)=0. 
$$
And, clearly,   the  following ``duality condition" holds true:
\begin{equation}\label{dual}
p\in  {\mathcal{H}}_{\bm\lambda}\Longleftrightarrow f_{\bm\lambda}(x_1(p), \ldots, x_n(p))=F(x_1(p), \ldots, x_n(p);\lambda_1,\ldots, \lambda_t)=0\Longleftrightarrow\bm\lambda\in \Gamma_p(\mathcal{F}).
\end{equation}

One may note that one classically refers to  the variety ${\bf I}\subset {\mathbb A}_{\x}^n(K)\times \mathcal{U}$  defined by the  polynomial $F({\x}; {\bm \Lambda})$  as  {\em incidence correspondence}, or {\em incidence variety}. It consists of the pairs of points
$(p, \bm \lambda)$ such that $p\in  {\mathcal{H}}_{\bm\lambda}$ or, equivalently,  $\bm\lambda\in \Gamma_p(\mathcal{F})$. In particular, denoting by ${\pi_1}_{|\bf I}:{\bf I}\to {\mathbb A}_{\x}^n(K)$, ${\pi_2}_{|\bf I}:{\bf I}\to \mathcal{U}$ 
the restrictions to ${\bf I}$  of  the product projections $\pi_1$, $\pi_2$ on the two factors, one has ${\pi_1}_{|{\bf I}}(\pi_2^{-1}(\bm\lambda))={\mathcal{H}}_{\bm\lambda}$ and, similarly,   ${\pi_2}_{|{\bf I}}(\pi_1^{-1}(p))=\Gamma_p(\mathcal{F})$ (see also \cite{BR}).

\smallskip

The following general facts hold true (see \cite[Theorem 2.2, Lemma 2.3]{BMP}, \cite[Section 3]{BR}).
 \begin{enumerate}
\item[1.]
{\em The Hough transforms ${\Gamma}_p(\mathcal F)$,  
when the point $p$ varies on ${\mathcal{H}}_{\bm\lambda}$, all pass through the point ${\bm\lambda}$}.
\item[2.] {\em Assume that  the Hough transforms ${\Gamma}_p(\mathcal F)$, 
when $p$ varies on ${\mathcal{H}}_{\bm\lambda}$,  have  a point in common other than ${\bm\lambda}$,
say ${\bm\lambda^{\prime}}$. Thus the two hypersurfaces ${\mathcal{H}}_{\bm\lambda}$, 
${\mathcal{H}}_{\bm\lambda^{\prime}}$ coincide}.
\item[3.]  (Regularity property)  {\em The following conditions are equivalent}:
\begin{enumerate}
\item[(a)] {\em For  any  hypersurfaces  ${\mathcal{H}}_{\bm\lambda}$,  ${\mathcal{H}}_{\bm\lambda^{\prime}}$ 
in $\mathcal{F}$, the equality  ${\mathcal{H}}_{\bm\lambda}={\mathcal{H}}_{\bm\lambda^{\prime}}$ 
implies ${\bm\lambda}={\bm\lambda^{\prime}}$}.
\item[(b)] {\em For each  hypersurface  $ {\mathcal{H}}_{\bm\lambda}$ in $\mathcal{F}$, one has
$\bigcap_{p\in {\mathcal{H}}_{\bm\lambda}}{\Gamma}_p(\mathcal F)=\{\bm\lambda\}$}.

\end{enumerate} 
\end{enumerate}

A family $\mathcal{F}$ which meets one of the above equivalent conditions (a), (b) is said to be
 {\em Hough regular}.

 \smallskip

From now on throughout the paper, we consider the case $n=2$. Let $\x=(x,y)$, and let  $\mathcal F=\{{\mathcal{C}}_{\bm\lambda}\}$  be a family of irreducible real curves in the image plane ${\mathbb A}_{(x,y)}^2(\R)$, of equation
\begin{equation}\label{GEcurve} 
f_{\bm\lambda}(x,y)=\sum_{i,j=0}^d x^iy^jg_{ij}({\bm\lambda}) , \;\;\;\; 0\leq i+j\leq d,
\end{equation} 
and satisfying   the assumptions and  the properties mentioned above (in particular, the ${\mathcal{C}}_{\bm\lambda}$'s  are  affine plane curves over $\C$ with infinitely many points in the affine plane over $\R$, see also \cite[Chapter 7]{SWPD}).

Given a profile of interest in the image plane, 
the Hough approach detects a curve from the family  $\mathcal F$ best approximating the profile
by using well-established pattern recognition techniques for the recognition of 
curves in images (see \cite[Sections 6, 7]{BMP} and also \cite[Sections 4, 5]{maetal}).
From a theoretical point of view,  the   detection procedure can be  highlighted as follows. 
\begin{enumerate}
\item[I.] Choose   a set
of points $p_j$'s of interest  in the image plane $ {\mathbb A}_{(x,y)}^2(\reals)$.
\item[II.] In the parameter space ${\mathbb A}_{\bm \Lambda}^t(\R)$ 
find the  intersection of the Hough transforms corresponding 
to the points $p_j$'s. That is, compute
$ \bigcap_j\Gamma_{p_j}(\mathcal F)$ which  identifies a (unique) point, ${\bm\lambda}$.\label{StepII}
\item[III.] Return the  curve ${\mathcal{C}}_{\bm\lambda}$ uniquely determined by the parameter ${\bm\lambda}$.
\end{enumerate}

Because of the presence of noise and approximations (due to the  floating point numbers representation encoding the real coordinates) of the points $p_j$'s extracted from a digital image, and consequently on their Hough transforms $\Gamma_{p_j}({\mathcal F})$, in practice in most cases it happens that  $\cap_j\Gamma_{p_j}({\mathcal F})= \emptyset$; though we notice that there are regions in the parameter space with high density of Hough transform crossings.
Therefore, from a practical point of view, Step II is usually performed using the 
so called ``voting procedure", a discretization  approach that  consists of the following three steps. 
\begin{itemize}
\item Find a proper discretization of a suitable bounded 
region ${\mathcal T}$ contained in the  open set  ${\mathcal{U}}\subset \R^t$
 of the parameter space. 
\item Construct on ${\mathcal T}$ an {\em accumulator function}, that is,  
 a function that, for each Hough transform  $\Gamma_{p_j}(\mathcal F)$ and for each cell 
of the discretized region, records and sums the ``vote" $1$, if $\Gamma_{p_j}(\mathcal F)$ 
crosses the cell, and the ``vote" $0$ otherwise.   
  \item  Look for the cell  associated to the maximum, say $\bf m$, of the accumulator function; 
 as suggested by the general results 
 recalled above, the center of that cell is an  approximation of the coordinates of the intersection point ${\bm\lambda}_{\bf m}$ (see \cite[Section 6]{BMP} and \cite[Section 4]{maetal}). 
Of course, such an approximation  is determined up to the chosen discretization of   ${\mathcal T}$.
\end{itemize}

Furthermore, in practice, Step I is performed by using a finite number of points of interest. Then it is natural to ask, even from a theoretical point of view, how many points $p_j$ are sufficient to uniquely identify $\bm \lambda$.

\subsection{Reduction to a finite intersection}\label{PS} Let ${\mathcal C}_{\bm \lambda}$ be a curve from the family 
${\mathcal F}=\{ {\mathcal C}_{\bm \lambda}\}$.
In general,  note that any (infinite) intersection 
${\mathscr T}:=\bigcap_{p\in{\mathcal C}_{ \bm \lambda}}\Gamma_{p}(\mathcal F)$
clearly reduces to a finite intersection of the same type. This simply because $K[\bm \Lambda]$ is a Noetherian ring
 (since $K={\mathbb R}$ or $K={\mathbb C}$ is Noetherian, it follows from the Hilbert basis theorem),  so that, since  every ascending chain of  ideals in $K[\bm\Lambda]$ is eventually stationary (e.g., see \cite{CA}), the ideal 
$I\subset K[\bm \Lambda]$
generated by the polynomials $f_p(\bm \Lambda)$  defining the Hough transforms $\Gamma_p({\mathcal F})$,
 $p\in {\mathcal C}_\lambda$, has a finite number, say $h$, of generators of ``Hough transform type" $f_p(\bm \Lambda)$ (and, clearly, a minimal finite number, say $m\leq h$, of generators  not necessarily of this type we are looking for).
 The natural question this raises is:
 \begin{quote} \item In the exact case, minimize the number of points $p_j$'s belonging to a curve ${\mathcal C}_{\bm \lambda}$ from the family
 ${\mathcal F}$ such that 
$$
{\mathscr T}=\bigcap_{p_j\in {\mathcal{C}}_{\bm\lambda},\; j\in \mathscr{J}}\Gamma_{p_j}(\mathcal F),$$
with $j$ belonging to a finite set $\mathscr{J}$ of  indices.  Coming to real applications,  this may    significantly reduce the time-consuming step of the recognition algorithm (see \cite[Section 4]{maetal} and also \cite[Section 5]{TB}), as shown  in Section \ref{App1}.  Clearly, ${\mathscr T}=\{\bm\lambda\}$   whenever the family ${\mathcal F}$ is  Hough regular. 
 \item 
  Forgetting about the regularity property, one may ask whether ${\mathcal C}_{\bm\lambda}={\mathcal C}_{\bm\lambda'}$ for any $\bm\lambda'\neq \bm\lambda$ belonging to ${\mathscr T}$.
For instance, in the case of families of real  plane curves,  we may ask if any point in ${\mathscr T}$ identifies the curve to be detected, this way extending the general fact II as in  the detection procedure highlighted  above (compare with Proposition \ref{A&Cprop}).
  \end{quote}

\section{A geometrical  bound}\label{FHT}
From now on throughout the paper,  we  consider the real case we are interested in. 
 Let ${\mathcal F}=\{{\mathcal{C}}_{\bm\lambda}\}$, $\bm\lambda\in {\mathcal{U}}\subseteq \R^t$, $t\geq 2$, be  a family of  real plane curves in ${\mathbb A}_{(x,y)}^2(\R)$ of equation (\ref{GEcurve}). By simplicity of notation, for families ${\mathcal{F}}$ of curves in  ${\mathbb A}_{(x,y)}^2(\reals)$ with $t=2,3$ parameters  we  set  $\bm\lambda=(a,b)$ and $\bm\lambda=(a,b,m)$, so that ${\mathbb A}_{(A,B)}^2(\reals)=\langle A,B\rangle$ and  ${\mathbb A}_{(A,B,M)}^3(\reals)=\langle A,B,M\rangle$ will denote the  parameter space, respectively.

 Consider  the projective closure of ${\mathcal{C}}_{\bm\lambda}:f_{\bm\lambda}(x,y)=0$ in the complex   projective plane $\pn 2(\C)$ of equation 
 $$\overline{{\mathcal{C}}_{\bm\lambda}}: f_{\bm\lambda}(x_0, x_1, x_2)=0,$$ where $f_{\bm\lambda}(x_0, x_1, x_2)\in \R[x_0,x_1,x_2]$ is the homogenization of $f_{\bm\lambda}(x,y)$  with respect to $x_2$, obtained by setting $x=\frac{x_0}{x_2}$, $y=\frac{x_1}{x_2}$.  Note that  $\overline{{\mathcal{C}}_{\bm\lambda}}$ is still an irreducible  curve of degree $d$, since
 $f_{\bm\lambda}(x_0, x_1, x_2)$ is irreducible over $\R$. We observe that the family $\overline{{\mathcal{F}}}=\{\overline{{\mathcal{C}}_{\bm\lambda}} \}$   is contained in a linear (or algebraic) system of curves.
 
  \begin{definition*}\label{defB} We say that  the set
 ${\mathcal{B}}={\mathcal{B}}(\C):=\{ p\in\pn 2(\C)\; |\; p\in  \overline{{\mathcal{C}}_{\bm\lambda}}\; \forall \bm\lambda\in {\mathcal{U}} \}$
is the {\em base locus} of the family $\overline{{\mathcal F}}=\{\overline{{\mathcal C}_{\bm\lambda}}\}$ of projective  curves in  $\pn 2(\C)$. 
We define
$${\mathcal{B}}_{\rm aff}=\{ p\in{\mathbb A}^2_{(x,y)}(\C)\; |\;p\in {\mathcal{C}}_{\bm\lambda}\; \forall \bm\lambda\in
 {\mathcal U} \}\;\;\;{\rm  and }\;\;\;
{\mathcal{B}}_\infty=\{ p\in\ell_\infty:x_2=0\; | \; p\in  \overline{{\mathcal C}_{\bm\lambda}}\; \forall \bm\lambda\in {\mathcal{U}} \},$$
where $\ell_\infty:x_2=0$ is  the line at infinity. 
\end{definition*}
 Clearly,
${\mathcal{B}}={\mathcal{B}}_{\rm aff}\cup {\mathcal{B}}_\infty$. We note that, under the irreducibility assumption on the curves from the family ${\mathcal{F}}$, both $ {\mathcal{B}}_{\rm aff}$ and $  {\mathcal{B}}_\infty$ consist of a finite number of points; we respectively denote by  $ \#{\mathcal{B}}_{\rm aff}$ and $ \#{\mathcal{B}}_\infty$  the number of points of such  sets. In particular, $\#{\mathcal{B}}=\#{\mathcal{B}}_{\rm aff}+  \#{\mathcal{B}}_\infty$.

As far as the  Hough  transform is concerned, note also that $\Gamma_p({\mathcal{F}})={\mathbb A}_{\bm\Lambda}^t(\R)$ for each real point $p\in {\mathcal{B}}_{\rm aff}$. Hence, in practical applications, one has to disregard the (real) points $p\in {\mathcal{B}}_{\rm aff}$.

Let us  point out the (although obvious) fact  that whenever $\Gamma_{p}(\mathcal F)=\Gamma_{q}(\mathcal F)$ for some points $p$, $q$ in the image space, then for each $\bm\lambda\in \Gamma_{p}(\mathcal F)$ the curve ${\mathcal{C}}_{\bm\lambda}\in {\mathcal{F}}$, which contains the point  $p$, has to pass through $q$ as well.\footnote{For practical purposes,  whenever $\Gamma_{p}(\mathcal F)=\Gamma_{q}(\mathcal F)$, then one of the two points $p$, $q$ is disregarded from the context.}

\smallskip

First, let us  consider the special (though relevant)   case when the parameters $\lambda_1,\ldots,\lambda_t$ linearly occur in equation $(\ref{GEcurve})$.

\begin{lemma}\label{linlem} Let ${\mathcal{F}}=\{{\mathcal{C}}_{\bm\lambda}\}$ be a family of real curves of degree $d$ in ${\mathbb A}_{(x,y)}^2(\R)$. 
Assume that  the polynomial expressions $g_{ij}(\bm\lambda)$ as in $(\ref{GEcurve})$ are linear in the  parameters $\lambda_1,\ldots,\lambda_t$.  Let ${\mathscr T}=\bigcap_{p\in{\mathcal C}_{\bm\lambda}}\Gamma_{p}(\mathcal F)$. Then the following conditions are equivalent:
\begin{enumerate}
\em\item\em For any curve ${\mathcal{C}}_{\bm\lambda}$ from the family  there exist  $t$  real points $p_j\in {\mathcal{C}}_{\bm\lambda}\setminus{\mathcal{B}}_{\rm aff}$
 such that the equations $f_{p_j}(\bm\Lambda)=0$ defining the Hough transforms  $\Gamma_{p_j}(\mathcal F)$ in the parameter space ${\mathbb A}_{\bm\Lambda}^t(\R)$ are linearly independent, $j=1,\ldots,t$.
\em\item\em  The family ${\mathcal F}$ is Hough regular  and ${\mathscr T} =\bigcap_{j=1}^t\Gamma_{p_j}(\mathcal F)=  
\{\bm\lambda\}$.
\end{enumerate}

\end{lemma}
\proof $1)\Rightarrow 2)$ The defining equations of the set $\bigcap_{j=1}^t\Gamma_{p_j}(\mathcal F)$ give rise to a linear system of $t$ equations in  $t$ variables $\Lambda_1,\ldots, \Lambda_t$, all of them vanishing at $\bm\lambda$. By the   assumption that the equations $f_{p_j}(\bm\Lambda)=0$, $j=1,\ldots,t$,  are linearly independent,  the rank of the matrix associated  to the  system  equals $t$, whence  ${\mathscr T}=\bigcap_{j=1}^t\Gamma_{p_j}(\mathcal F)=\{\bm\lambda\}$,  so that ${\mathcal F}$ is Hough regular by the equivalent condition (b) of the regularity property. 

$2)\Rightarrow 1)$ Arguing by contradiction, assume that there exists a curve ${\mathcal{C}}_{\bm\lambda}$ from the family such that for any $t$ points $p_j\in {\mathcal{C}}_{\bm\lambda}\setminus{\mathcal{B}}_{\rm aff}$  the equations $f_{p_j}(\bm\Lambda)=0$ defining the Hough transforms  $\Gamma_{p_j}(\mathcal F)$, $j=1,\ldots,t$,  are linearly dependent. Then they 
give rise  to a linear system of $t$ equations in  $t$ variables $\Lambda_1,\ldots, \Lambda_t$ with  infinitely many real solutions. Let $\bm\lambda'\neq \bm\lambda$ one of them. By   duality condition (\ref{dual}) it then follows that $p_1,\ldots,p_t\in 
 {\mathcal C}_{\bm\lambda'}$, whence    ${\mathcal C}_{\bm\lambda}\subseteq {\mathcal C}_{\bm\lambda'}$. Thus, passing to the projective closures,  one has   $\overline{{\mathcal{C}}_{\bm\lambda}}=\overline{{\mathcal{C}}_{\bm\lambda'}}$.  Therefore, restricting to the affine plane ${\mathbb A}_{(x,y)}^2(\C)=\pn 2(\C)\setminus \ell_\infty$,
  it must be ${\mathcal{C}}_{\bm\lambda}={\mathcal{C}}_{\bm\lambda'}$ in ${\mathbb A}_{(x,y)}^2(\C)$, whence ${\mathcal{C}}_{\bm\lambda}={\mathcal{C}}_{\bm\lambda'}$ in ${\mathbb A}_{(x,y)}^2(\R)$ since the two curves are real. This contradicts the Hough  regularity assumption.
\qed
\smallskip

The following example shows that the assumption on the defining  equations to be linearly independent in  statement $1)$  of the above  lemma is  needed.

\begin{example*}\label{banale}
Consider the family ${\mathcal{F}}=\{{\mathcal C}_{a,b}\}$ of cubic curves of equation
$${\mathcal{C}}_{a,b}:y^2=x^3+ax + b,$$ for  real parameters $\bm\lambda=(a,b)$. Take the cubic  ${\mathcal{C}}_{\bm\lambda}:y^2= x^3 + x+1$, $\bm\lambda=(1,1)$, and the points $p_1=(0,1)$, $p_2=(0,-1)$ on ${\mathcal{C}}_{1,1}$.
Then $\Gamma_{p_1}({\mathcal{F}})=\Gamma_{p_2}({\mathcal{F}}):B=1$,  so that  the set $\Gamma_{p_1}({\mathcal F})\cap\Gamma_{p_2}({\mathcal{F}})$  coincides  with the line $B=1$ in the parameter plane $\langle A, B\rangle$. Moreover, 
${\mathcal{C}}_{1,1}\neq {\mathcal{C}}_{a,1}$ for $a\neq 1$, this showing that  given  $\bm\lambda, \bm\lambda'\in \Gamma_{p_1}({\mathcal F})\cap\Gamma_{p_2}({\mathcal{F}})$ does not follow that ${\mathcal{C}}_{\bm\lambda}={\mathcal{C}}_{\bm\lambda'}$.
\quadrato\end{example*}

Let us consider now  the general case. A simple geometrical argument, based on  B\'ezout theorem, leads to a natural   finite bound.  Even though it is not sharp,   as the examples in Section \ref{EX} show, it looks of interest  for practical purposes (see Section \ref{App1}, and also   \cite[Section 2]{HTbone}).

\begin{prop}\label{A&Cprop} 
Let ${\mathcal{F}}=\{{\mathcal{C}}_{\bm\lambda}\}$ be a family of real curves of degree $d$ in 
${\mathbb A}_{(x,y)}^2(\R)$.  Let ${\mathcal{B}}(\C)$ be the base locus of the associated  family ${\overline{\mathcal F}}=\{\overline{{\mathcal{C}}_{\bm\lambda}} \}$ of  projective curves in  $\pn 2(\C)$, and 
set  $\nu_{\rm opt}:=d^2-\#{\mathcal{B}}(\C)+1$. For any  curve ${\mathcal{C}}_{\bm\lambda}$ from the family take $\nu_{\rm opt}$ arbitrarily chosen real distinct points $p_j\in{\mathcal{C}}_{\bm\lambda}\setminus{\mathcal{B}}_{\rm aff}$, $j=1,\ldots,\nu_{\rm opt}$.
Let ${\mathscr T}=\bigcap_{p\in{\mathcal C}_{\bm\lambda}}\Gamma_{p}(\mathcal F)$, and set
${\mathscr T}_{\rm opt}:=\bigcap_{j=1,\ldots, \nu_{\rm opt}}\Gamma_{p_j}(\mathcal F)$. 
Then: 
\begin{enumerate}
\em\item\em ${\mathcal{C}}_{\bm\lambda '}={\mathcal{C}}_{\bm\lambda}$ for each   $\bm\lambda' \in {\mathscr T}_{\rm opt}$.
\em\item\em ${\mathscr T}_{\rm opt}={\mathscr T}$.
\em\item\em  If the family ${\mathcal{F}}$  is  Hough regular,     
then ${\mathscr T}_{\rm opt}=\{\bm\lambda\}$.
\end{enumerate}
\end{prop}

\proof
 For a given (real) point $\bm\lambda'\in {\mathscr T}_{\rm opt}$, consider the curves ${\mathcal{C}}_{\bm\lambda}$, ${\mathcal{C}}_{\bm\lambda'}$. Since $\Gamma_{p_j}({\mathcal{F}})\ni \bm\lambda'$,  duality condition (\ref{dual}) assures that  ${\mathcal{C}}_{\bm\lambda'}\ni p_j$, $j=1,\ldots, \nu_{\rm opt}$. It thus follows that  the projective closure curves  $\overline{{\mathcal{C}}_{\bm\lambda}}$,  $\overline{{\mathcal{C}}_{\bm\lambda'}}$ in the complex projective plane
 $\pn 2(\C)$ (which have in common the $\#{\mathcal{B}}(\C)$ points of the base set ${\mathcal{B}}:={\mathcal{B}}(\C)$) meet in at least 
$$\nu_{\rm opt} +\#{\mathcal{B}}(\C)=d^2-\#{\mathcal{B}}(\C)+1 +\#{\mathcal{B}}(\C)=d^2+1$$ (distinct)  points of $\pn 2(\C)$.
On the other hand,   the assumptions that the  family ${\mathcal{F}}$ consists of irreducible curves sharing the degree implies that  the curves $\overline{{\mathcal{C}}_{\bm\lambda}}$,  $\overline{{\mathcal{C}}_{\bm\lambda'}}$  don't have common components. Thus, 
B\'ezout's theorem (see e.g. \cite[\S 4.2]{BCGM})  allows us to conclude that
$\overline{{\mathcal{C}}_{\bm\lambda}}=\overline{{\mathcal{C}}_{\bm\lambda'}}$.  Therefore, restricting to the affine plane ${\mathbb A}_{(x,y)}^2(\C)=\pn 2(\C)\setminus \ell_\infty$,  it must be ${\mathcal{C}}_{\bm\lambda}={\mathcal{C}}_{\bm\lambda'}$ in ${\mathbb A}_{(x,y)}^2(\C)$, whence ${\mathcal{C}}_{\bm\lambda}={\mathcal{C}}_{\bm\lambda'}$ in ${\mathbb A}_{(x,y)}^2(\R)$ since the two curves are real.
This proves  the first assertion. 

In order to prove the second assertion, we only have to prove the inclusion 
${\mathscr T}_{\rm opt} \subseteq {\mathscr T}$. If ${\mathscr T}_{\rm opt}= \{\bm\lambda\}$ this is clear, since $\bm\lambda\in {\mathscr T}$
by duality condition~(\ref{dual}).
Now, let's consider  the case  ${\mathscr T}_{\rm opt} \neq \{\bm\lambda\}$.  
By contradiction, we assume that there exists $\bm\lambda'  \in {\mathscr T}_{\rm opt}$, with $\bm\lambda' \neq \bm\lambda$, 
such that $\bm\lambda'  \not\in {\mathscr T}$. Therefore, there exists a point $q \in {\mathcal C}_{\bm\lambda}$
such that $\bm\lambda' \not\in \Gamma_q(\mathcal F)$. By duality condition~(\ref{dual}) this is equivalent to say that 
$q \not\in {\mathcal C}_{\bm\lambda'}$, contradicting assertion  $1)$.

Finally, assuming Hough regularity for the family ${\mathcal{F}}$, it then follows  $\bm\lambda=\bm\lambda'$, whence
 ${\mathscr T}_{\rm opt}=\{\bm\lambda\}$, which completes the proof.
\qed

\begin{example*}\label{Ex2}
Consider  in ${\mathbb A}^2_{(x,y)}(\mathbb R)$ the family ${\mathcal{F}}=\{{\mathcal{C}}_{a,b}\}$ of conics of equation
$$ a(x^2+y^2+1)+b(x^2+x+y)=0,$$
for  real parameters $\bm\lambda=(a,b)$. Since $a$, $b$ are defined up to a non-zero constant, the family ${\mathcal{F}}\cong\pn 1_{[a,b]}(\R)$ is in fact a pencil of conics.  We then have $\#{\mathcal{B}}(\C)=4$, so that $\nu_{\rm opt}=1$. 
This means that, for each single point $p$  taken on a fixed conic ${\mathcal{C}}_{\bm\lambda}$ of the pencil, one has
$${\mathscr T}_{{\rm opt}}= \Gamma_p({\mathcal{F}}) :(x_p^2+y_p^2+1)A+(x_p^2+x_p+y_p)B=0.$$ 
Therefore, by Proposition \ref{A&Cprop}(1), for any $\bm\lambda'$  belonging to the  line $\Gamma_p({\mathcal{F}})$
one has ${\mathcal{C}}_{\bm\lambda}={\mathcal{C}}_{\bm\lambda'}$. This agrees with the fact that  the family ${\mathcal{F}}$ is clearly  not Hough regular, since 
${\mathcal{C}}_{a,b}={\mathcal{C}}_{ka,kb}$ for each $k\in \R^*$.\quadrato
\end{example*}

\section{Examples  of interest}\label{EX}

In this section we provide the  examples   we come back on in next Section \ref{App1}. Such examples belong to  classes of curves of interest in astronomical and medical imaging, 
 and widely used in recent literature to best approximate bone profiles and typical solar structures such as coronal loops (for instance, see \cite{BMP, maetal, ICIAP}). These families of curves mainly come from atlas of plane curves as \cite{atlas}, as well as from knowledge of classical tools in algebraic geometry.

We use the  notation as in the previous sections.  Moreover,  for  a point $p=(x_p,y_p)$ in the image plane ${\mathbb A}_{(x,y)}^2(\reals)$, we denote by $[x_0(p), x_1(p),x_2(p)]$ its  homogeneous coordinates in the real  projective plane $\pn 2_{[x_0,x_1,x_2]}$.

\begin{example*}(Descartes Folium)\label{FC}
Consider  the family ${\mathcal F}=\{\mathcal{C}_{a,b}\}$ of cubic rational curves defined by the equation
\begin{equation}\label{DF} \mathcal{C}_{a,b}: 3axy-x^3-by^3=0,\end{equation} for some real parameters $a$, $b$ such that $ab\neq 0$ (for $b=1$, such a cubic is classically known as the {\em Descartes Folium}). 
Such a curve has  a node at the origin and a loop in the first (respectively, second) quadrant  if $b>0$ (respectively, $b<0$) (see also \cite[Section 3]{BMP}  for a more detailed description).

Passing to homogeneous coordinates we have
$$\overline{{\mathcal C}_{a,b}}: 3ax_0x_1x_2-x_0^3-bx_1^3=0.$$
The base locus ${\mathcal{B}}(\C)$ of the family $\overline{{\mathcal F}}=\{
\overline{{\mathcal C}_{a,b}} \} $ consists of the points $p\in \pn 2(\C)$ such that the polynomial
$$3ax_0(p)x_1(p)x_2(p)-x_0(p)^3-bx_1(p)^3$$
is identically zero  in  $ \R[a,b]$. Then $p=[x_0,x_1,x_2]\in {\mathcal{B}}(\C)$ if and only if it is a solution of the system
$$x_0x_1x_2=x_0^3=x_1^3=0,$$ so that
${\mathcal{B}}(\C)=\{[0,0,1]\} $. Therefore the bound from Proposition \ref{A&Cprop}   becomes
$$\nu_{\rm opt}=d^2-\#{\mathcal{B}}(\C)+1=9-1+1=9.$$

On the other hand, according to Lemma \ref{linlem}, for any pair of points $p_1,p_2\in \mathcal{C}_{\bm\lambda}$, $\bm\lambda=(a,b)$, one  has in fact
$$\Gamma_{p_1}({\mathcal{F}})\cap\Gamma_{p_2}({\mathcal{F}})=\{\bm\lambda\}$$  as soon as the equations $f_{p_1}(A,B)$, $f_{p_2}(A,B)$ are linearly independent.
\end{example*}

\begin{example*}(Elliptic curves)\label{CE}
Consider the family  $\mathcal{F}=\{ {\mathcal C}_{a,b,m} \}$ of unbounded cubic curves of equation
\begin{equation}\label{ell1}
 {\mathcal C}_{a,b,m}: y^2=mx^3+ax+b,
\end{equation}
for  non-zero real parameters $a$, $b$, $m$. Non-singular curves from the family have genus $1$ and are called {\em elliptic curves}. For $m=1$,  one refers to equation (\ref{ell1}) as the {\em Weierstrass equation} of the curve (see \cite[\S 3.2]{maetal}).

For any point $p=(x_p,y_p)$, the Hough transform is the plane $\Gamma_p({\mathcal F})$, in the parameter space ${\mathbb A}_{(A,B,M)}^3(\reals)$, of equation
$$f_p(A,B,M) : x_pA+B+x_p^3M-y_p^2=0.$$

Let $\bm\lambda=(1,1,1)$ and take the points $p_1=(0,1)$, $p_2=\big(\frac{1}{2}, \sqrt{\frac{13}{8}}\big)$ on the curve ${\mathcal{C}}_{1,1,1}$. Then $\Gamma_{p_1}({\mathcal{F}})\cap \Gamma_{p_2}({\mathcal{F}})$ is the line $\ell:B-1=4A+M-5=0$ in $\langle A,B,M\rangle$, which contains the point $(1,1,1)$. Among the curves ${\mathcal{C}}_{a,1,5-4a}$ corresponding to the point $(a,1,5-4a)\in \ell$, choose  for instance  ${\mathcal{C}}_{\frac{1}{4},1,4}$. One then sees that ${\mathcal{C}}_{1,1,1}\neq{\mathcal{C}}_{\frac{1}{4},1,4}$ and ${\mathcal{C}}_{1,1,1}$, ${\mathcal{C}}_{\frac{1}{4},1,4}$ meet in exactly six  points (pairwise symmetric to the $x$-axis) in the image plane ${\mathbb A}_{(x,y)}^2(\reals)$.

According to Lemma \ref{linlem}, as soon as one takes a third   point $p_3$ on ${\mathcal{C}}_{1,1,1}$ such that  the equations $f_{p_i}(A,B,M)=0$, $i=1,2,3$, are linearly independent (in particular, $p_3\neq (0,-1)$ since $\Gamma_{(0, -1)}({\mathcal{F}})=\Gamma_{p_1}({\mathcal{F}})$) one gets
$$\Gamma_{p_1}({\mathcal{F}})\cap\Gamma_{p_2}({\mathcal{F}})\cap \Gamma_{p_3}({\mathcal{F}})=\{\bm\lambda\}.$$ 
\end{example*}

\begin{example*} (Quartic curve with a triple point)\label{Q3}
Consider the family $\mathcal{F}=\{ {\mathcal C}_{a,b}  \}$ of quartic  curves defined by the equation
\begin{equation}\label{Q3eq}
{\mathcal C}_{a,b} : y(x-ay)^2-b(x^2+y^2)^2=0,
\end{equation}
for real parameters $a$, $b$ with $b>0$. The curve ${\mathcal{C}}_{a,b}$ has a triple point at the origin, so  it is a rational curve. 
As to the variance, the curve ${\mathcal C}_{a,b} $ is contained  in the semi-circumference with center $ (0,0)$  and radius  $R_{a,b}=\dfrac{(1+|a|)^2}{b}$
(see \cite[\S 4.1 and Section 7]{BMP}).
Passing to homogeneous coordinates we have
$$\overline{{\mathcal{C}}_{a,b}}: x_1(x_0-ax_1)^2x_2-b(x_0^2+x_1^2)^2=0.$$
The base locus ${\mathcal{B}}(\C)$ of the family $\overline{{\mathcal{F}}}=\{ \overline{{\mathcal{C}}_{a,b}}\} $ consists of the points $p\in \pn 2$ such that 
$$a^2x_1(p)^3x_2(p)-2ax_0(p)x_1^2(p)x_2(p)-b(x_0(p)^2+x_1(p)^2)^2+x_1(p)x_0(p)^2x_2(p)$$
is an identically zero polynomial in $ \R[a,b]$. Then $p=[x_0,x_1,x_2]\in {\mathcal{B}}(\C)$ if and only if it is a solution of the system
$$x_1^2x_2=x_0x_1^2x_2=(x_0^2+x_1^2)^2=x_1x_0^2x_2=0,$$ so that
${\mathcal{B}}(\C)=\{[0,0,1], [\pm i, 1,0]\}$. Therefore the bound from Proposition \ref{A&Cprop}   becomes
$$\nu_{\rm opt}=d^2-\#{\mathcal{B}}(\C)+1=16-3+1=14.$$

\end{example*}

\begin{example*} (Quartic curve with a tacnode)\label{Qtac}
Consider the family $\mathcal{F}$ of quartic  curves defined by the equation
\begin{equation}\label{Qtaceq}
{\mathcal{C}}_{a,b} : y^2(x-a)^2-byx^2+x^4=0,
\end{equation}
for real parameters $a$, $b$ with $b>0$. The curve ${\mathcal{C}}_{a,b}$ has a cusp at the origin $O$,  with cuspidal tangent the line $\ell:y=0$  and intersection multiplicity $m_O(\ell,{\mathcal{C}}_{a,b})=4$ (such a singularity is called a {\em tacnode}) and one more singular point at the infinity, so  it is a rational curve.

The real points of such curves present a single closed loop and a loop closed at the infinity (see \cite[\S 4.3 and Section 7]{BMP}).
Passing to homogeneous coordinates we have
$$\overline{{\mathcal{C}}_{a,b}}: x_1^2(x_0-ax_2)^2-bx_1x_0^2x_2+x_0^4=0.$$
The base locus ${\mathcal{B}}(\C)$ of the family $\overline{{\mathcal{F}}}$ consists of the points $p\in \pn 2$ such that the polynomial
$$a^2x_1(p)^2x_2(p)^2-2ax_0(p)x_1(p)^2x_2(p)-bx_0(p)^2x_1(p)x_2(p)+x_0(p)^4+x_0(p)^2x_1(p)^2=0 $$
is identically zero in $ \R[a,b]$. Then $p=[x_0,x_1,x_2]\in {\mathcal{B}}(\C)$ if and only if it is a solution of the system
$$x_1^3x_2^2=x_0x_1^2x_2=x_0^2x_1x_2=x_0^4+x_0^2x_1^2=0,$$ so that
${\mathcal{B}}(\C)=\{[0,0,1], [0,1,0],[\pm i, 1,0]\}$. Therefore the  bound from Proposition \ref{A&Cprop}   becomes
$$\nu_{\rm opt}=d^2-\#{\mathcal{B}}({\mathcal{C}})+1=16-4+1=13.$$
\end{example*}

\section{Applications to synthetic data}\label{App1}

In this section we show the efficiency of the   bound discussed in Section \ref{FHT} for four  families of curves considered in \cite{BMP}. In particular,  we  show the robustness of the  results when applied to dataset strongly perturbed by noise. We keep the same notation as in  \cite[Section 6]{BMP}.

From now on, we consider the following set of curves selected from four families: 
the Descartes Folium of equation (\ref{DF}) with $a= 3$, $b=1$, the elliptic curve of equation (\ref{ell1}) with $a= -4 $, $b=7$, the quartic curve with triple point  of equation (\ref{Q3eq}) with $a= \frac{1}{5} $, $b=\frac{1}{2}$;  and the quartic curve with  tacnode of equation (\ref{Qtaceq}) with $a=1  $, $b=8$.  These are exactly the same curves considered in  \cite[Section 6]{BMP}; from now on, we also refer to them as ``the given curves".
As stated in  Section \ref{StepII}, for a successful recognition of the given curves, we need to find the intersection of the Hough transforms in order to identify ${\bm\lambda}.$ The voting procedure requires some steps: first of all, we need to bound the parameter space, selecting minimum and maximum values for the parameters  to be considered. In the following we indicate these values with  $a_{{\rm min}}$, $a_{{\rm max}}$, $b_{{\rm min}}$, and $b_{{\rm max}}$ for $A$ and $B$, respectively. Then, we discretize the region in the  parameter space, choosing the cell size $\delta_{a}$ along the  $A$ axis, and the cell size $\delta_{b}$ along the  $B$ axis.  The number of cells along  the $A$ axis of the parameter space is  then computed as: 
$$N_a = \lfloor{\frac{a_{{\rm max}}-a_{{\rm min}}}{\delta_a}}\rfloor,  $$
and  an analogous formula  holds for $N_b.$

All the values considered in the four cases are collected in Table \ref{T0}. The  parameter spaces are built in such a way that each of them contains a cell corresponding to the  pair $(a,b)$ employed to select the curves. In this way, we can achieve an exact recognition, where the error between the original parameters and the recognized ones is equal to zero. Further, it is worth noting that to make the comparison with the results presented in \cite{BMP} more reliable, in the four cases under consideration we have sampled the same regions of the $\langle x, y \rangle$  plane and considered the same discretizations of the parameter spaces, as previously done in \cite{BMP}.

\begin{table}[h]
\begin{center}
\begin{tabular}{c|c|c|c|c||c|c|c|c|}
\cline{2-9}
& \multicolumn{4}{|c||}{$A$} & \multicolumn{4}{|c|}{$B$} \\
\hline
\multicolumn{1}{ |c | }
{{\em Family of curves}} & $a_{{\rm min}} $ & $a_{{\rm max}} $ & $\delta_{a} $ & $N_a$  & $b_{{\rm min}} $ & $b_{{\rm max}} $ & $\delta_{b} $  & $N_b$\\
 \hline  \hline
\multicolumn{1}{ |c | }{Descartes Folium} & $0.5$ & $11$ & $0.02$ & $525$ & $0.5$ & $11$ & $0.02$ & $525$\\
 \hline 
\multicolumn{1}{ |c | }{Elliptic curve} &  $-14$ & $6$ & $0.02$ & $1000$ &  $-3$ & $17$ & $0.02$ & $1000$\\
 \hline 
\multicolumn{1}{ |c | }{Quartic curve with triple point} &  $-5$ & $5$ & $0.01$ & $1000$ &  $0.1$ & $5$ & $0.01$ & $490$\\
 \hline 
\multicolumn{1}{ |c | }{Quartic curve with tacnode} &  $-9$ & $11$ & $0.02$ & $1000$ &  $-2$ & $18$ & $0.02$ & $1000$\\
\hline
\end{tabular}
\end{center}
\caption{\small{Values used to discretize the parameter space for the four families of curves.}}
\label{T0}
\end{table}

\subsection{Robustness in absence of noise}\label{NF}
\begin{figure}[h]
\begin{center}
\begin{tabular}{cc}
\includegraphics[height=4cm]{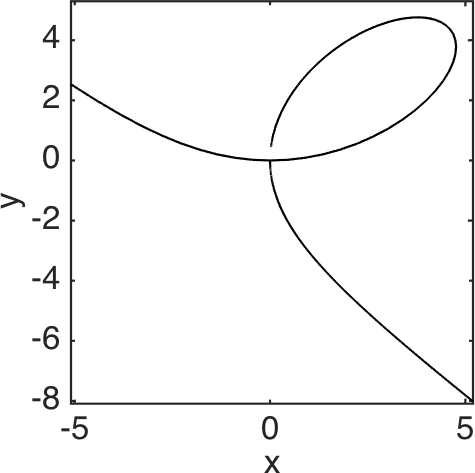} &\includegraphics[height=4cm]{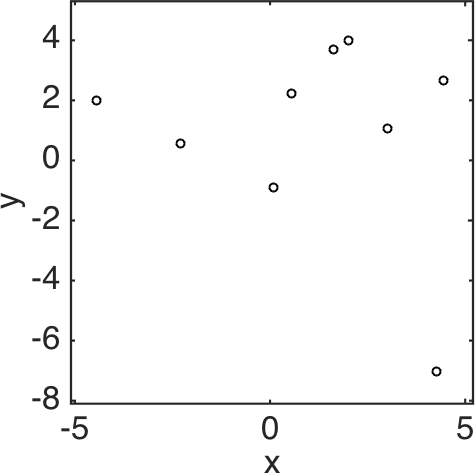} \\
\hspace{0.5cm} (a) & \hspace{0.5cm} (b) \\
\includegraphics[height=4cm]{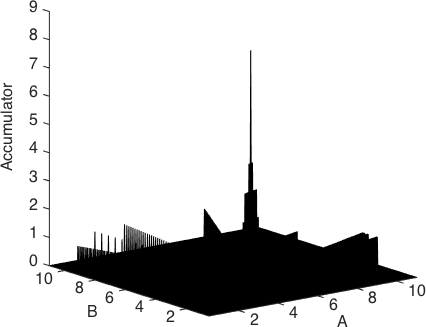} &\includegraphics[height=4cm]{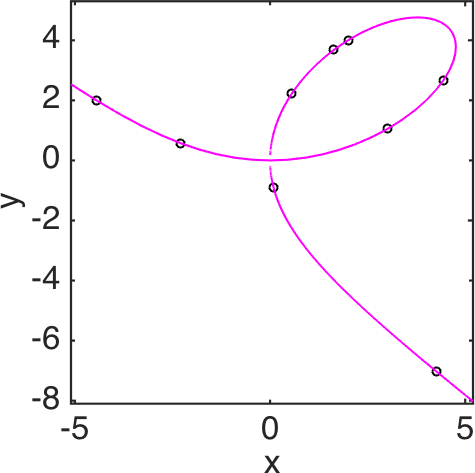} \\
\hspace{0.5cm} (c) & \hspace{0.5cm} (d)
\end{tabular}
\caption{\small{Recognition of the Descartes folium using $\nu_{\rm opt} = 9$ points. The curve given by $a = 3$, $b = 1$, panel (a); $\nu_{\rm opt}$ points  randomly sampled on
the curve, panel (b); the accumulator function,  panel (c); the recognized curve, in magenta, with sampled points superimposed, panel (d).}}
\label{F0}
\end{center}
\end{figure}

We start the analysis testing the bounds  given in Proposition \ref{A&Cprop} when no noise is present: for each curve described above we randomly select $\nu_{\rm opt}$ points and apply the recognition algorithm. We repeat the  random extraction procedure for $100$ runs, in order to assess the robustness with respect to the choice of the points  in the dataset. For  the whole set of curves, we recognize the exact pair of parameters in all the runs. In the first row of  Figure \ref{F0} we show  as an example the curve of the Descartes Folium family with $a=3$, $b=1$ (panel (a)), and $\nu_{\rm opt} = 9$ points randomly sampled from it (black  circles) (panel (b)); in panel (c) of Figure \ref{F0}  we present the accumulator function that has a clear peak in  the cell corresponding to $(a,b)=(3,1)$. This cell is selected as the one  corresponding to the maximum value of the accumulator and provides  us with the parameters of the reconstructed curve (see panel (d)). 

\subsection{Robustness in presence of noisy background}\label{RNB}

In this paragraph we present the results  concerning the robustness in presence of  a very noisy  background around the selected curves as above.
For each curve  we build a database made of $N_1 + N_2$ points where   $N_1 = \nu_{\rm opt}$ points (dataset points, from now on)  satisfy the curve equation, and $N_2$ points  (noise points, from now on) are randomly picked up on the image plane according to a uniform distribution. We test different levels of background noise ($99\%$, $95\%$, $90\%$, $85\%$, $80\%$), considering 
\begin{equation}\label{formula}
\frac{N_2}{\nu_{\rm opt}+N_2}=\frac{x}{100},
\end{equation}
where $x = 99, 95, 90, 85, 80$. For the four given curves, the values of $N_1$ and $N_2$ and the total number $N = N_1 + N_2$ of Hough transforms, which  depends on the background noise level,  are summarized in Table \ref{T1}.
 Let us remark that  the quantity $N$ is 
 definitely lower than the corresponding one employed in  \cite{BMP}. For instance, in the case of background noise at $99\%$ (the only case made explicit in \cite{BMP}), here we employ $900$ points for the Descartes Folium and the elliptic curve, $1400$ for the quartic curve with triple point, and $1300$ for the quartic curve with tacnode, versus  $10000$ for the Descartes Folium, $15800$ for the elliptic curve, $10000$ for the quartic curve with triple point, and $5000$ for the quartic curve with tacnode as reported in Table \ref{T1} of   \cite{BMP}.

\begin{table}[h]
\begin{center}
\begin{tabular}{cc|c|c||c|c||c|c||c|c||c|c|}
\cline{3-12}
& & \multicolumn{10}{ c| }{{\em Noise level}} \\ 
\cline{3-12}
&& \multicolumn{2}{|c||}{$99\%$}& \multicolumn{2}{|c||}{$95\%$} & \multicolumn{2}{|c||}{$90\%$} &\multicolumn{2}{|c||}{$85\%$} & \multicolumn{2}{|c|}{$80\%$}\\
 
\hline
\multicolumn{1}{ |c | }
{{\em Family of curves}} & 
\multicolumn{1}{ c||}{$N_1$} & $N_2$ & $N$  & $N_2$ & $N$  & $N_2$ & $N$  & $N_2$ & $N$  & $N_2$ & $N$ \\
 \hline  \hline
\multicolumn{1}{ |c | }{Descartes Folium} & \multicolumn{1}{ c||}{$9$} & $891$ &  $ 900$ &  $171$ & $180$ & $81$ &$90$ & $51$ & $60$ & $36$ & $45$\\
 \hline 
\multicolumn{1}{ |c | }{Elliptic curve} & \multicolumn{1}{ c||}{$9$} & $891$ &  $ 900$ &  $171$ & $180$ & $81$ &$90$ & $51$ & $60$ & $36$ & $45$\\
  \hline 
\multicolumn{1}{ |c | }{Quartic curve with triple point} & \multicolumn{1}{ c||}{$14$}& $1386$ & $1400$& $266$ & $280$& $126$ &$140$& $80$ &$94$ & $56$ &$70$ \\
  \hline 
\multicolumn{1}{ |c | }{Quartic curve with tacnode} & \multicolumn{1}{ c||}{$13$} & $1287$ &$1300$& $247$ &$260$ & $117$ &$130$ & $74$ & $87$& $52$ &$65$\\
  \hline 
\end{tabular}
\end{center}
\caption{\small{Number of points used in the robustness test for the Hough transform recognition method:
$N_1=\nu_{\rm opt}$, the  number of points on each curve, $N_2$, the  number of background noise points satisfying the
condition $\frac{N_2}{\nu_{\rm opt}+N_2}=\frac{x}{100}$ in the case of noise $\frac{x}{100}=99\%,
95\%, 90\%, 85\%, 80\%$, and $N = N_1 + N_2$, the total number of points.}}
\label{T1}\end{table}

We repeated the experiments for $ 100$ runs, randomly extracting the $N_1$ points on the curve and the $N_2$ background noise points. In Table \ref{T2}  we show the number of runs (out of $100$) in which the method correctly recognizes the parameters, while in Table \ref{T3} we show the average distance and  the corresponding standard deviation  between the pair of exact parameters and the recognized ones.

\begin{figure}[h]
\begin{center}
\begin{tabular}{cccc}
\includegraphics[scale=0.5]{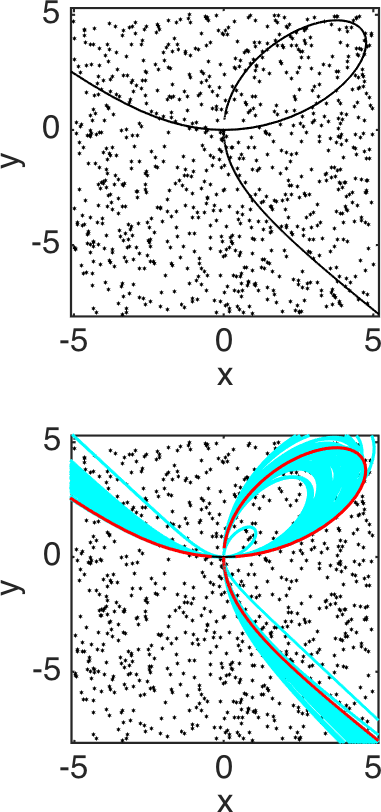} &
\includegraphics[scale=0.5]{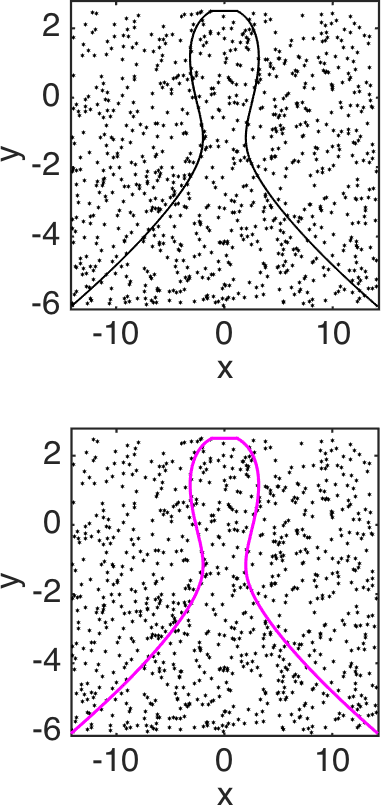}&
\includegraphics[scale=0.5]{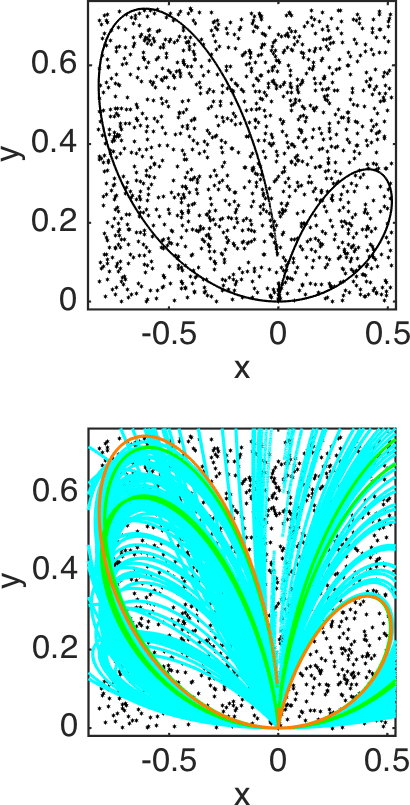}&
\includegraphics[scale=0.5]{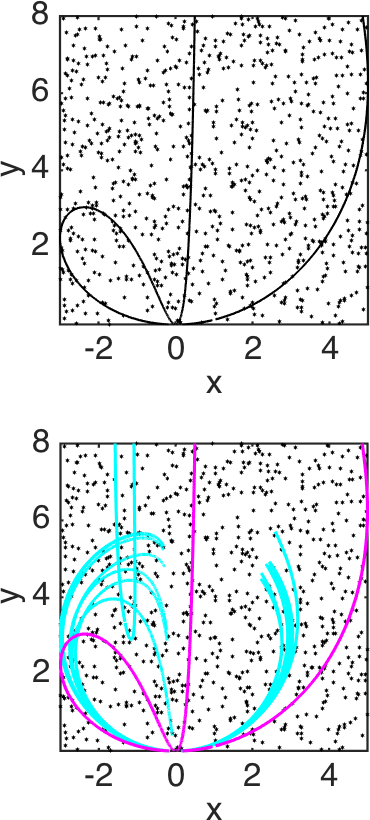}\\
\hspace{0.5cm} (a) & \hspace{0.5cm} (b) & \hspace{0.5cm} (c) & \hspace{0.5cm} (d)
\end{tabular}
\caption{\small{Recognition of the Descartes Folium, panel (a), the elliptic curve, panel (b),  the quartic  curve with a triple point, panel (c), and the quartic curve with tacnode, panel (d),  when embedded in a noisy background ($99\%$ of background noise points). For each case,  and for a run out of the $100$ we considered, the first row represents the noise points (dots) and the given curve (solid), whereas the second row represents the noise points (dots) and the recognized curves (solid). The colors of the curves are associated to their repetition rates.}}
\label{F1}
\end{center}
\end{figure}

\begin{figure}[h]
\begin{center}
\begin{tabular}{cccc}
\includegraphics[scale=0.5]{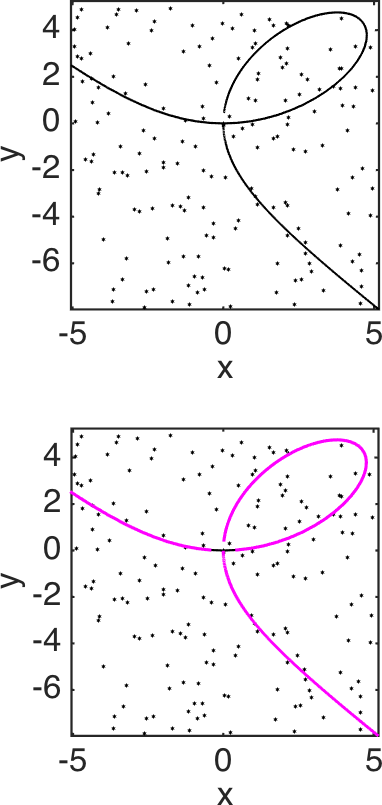} &
\includegraphics[scale=0.5]{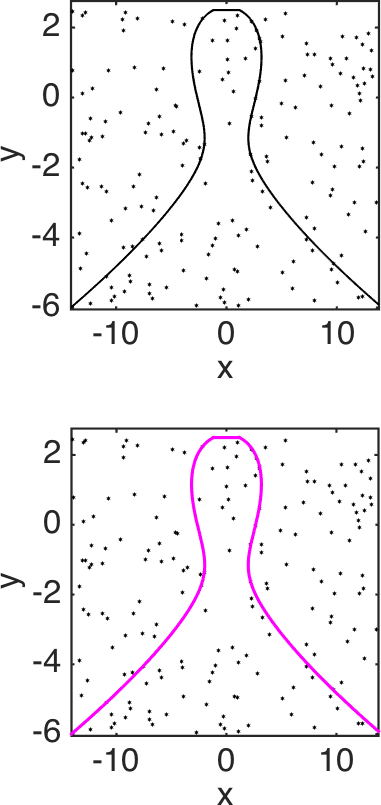} & 
\includegraphics[scale=0.5]{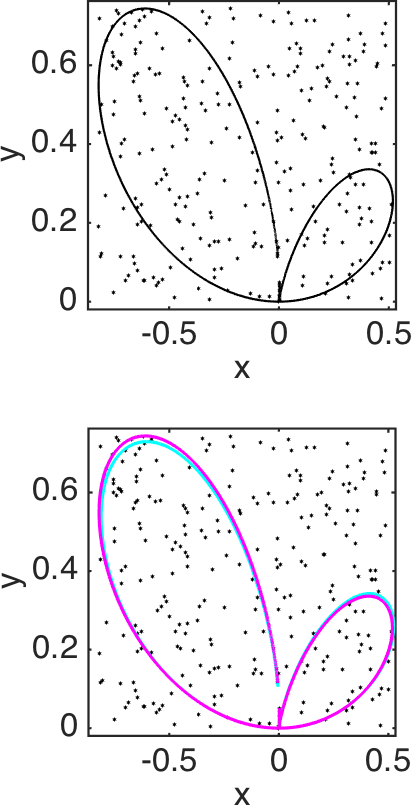} &
\includegraphics[scale=0.5]{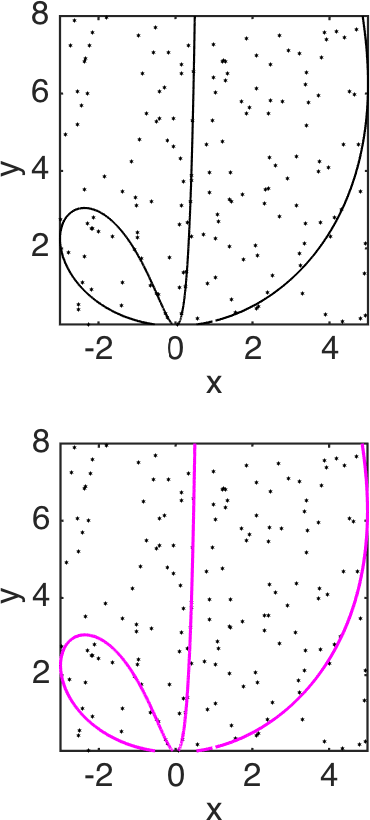}\\
\hspace{0.5cm} (a) & \hspace{0.5cm} (b) & \hspace{0.5cm} (c) & \hspace{0.5cm} (d)
\end{tabular}
\caption{\small{Recognition of the Descartes Folium, panel  (a), the elliptic curve, panel  (b), the quartic  curve with a triple point, panel  (c), and the quartic curve with tacnode, panel  (d), when embedded in a noisy background ($95\%$ of background noise points).   For a run out of the $100$ we considered, the first row in each panel represents the noise points (dots), and the given curve (solid), whereas the second row represents the noise points (dots) and the recognized  curves (solid). The  magenta color of the curves means that   their repetition rates are  higher than $50\%$; in the case of  the quartic with a triple point a cyan colored curve occurs as well (repetition rate from 2\% to 3\%).}}
\label{F2}
\end{center}
\end{figure}

\begin{figure}[h]
\begin{center}
\includegraphics[scale=0.5]{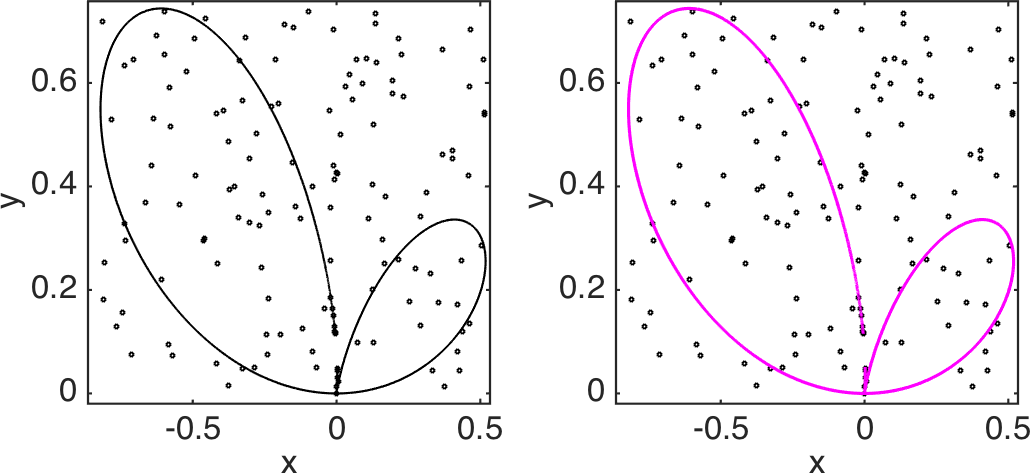}\\
\caption{\small{Recognition of   the quartic  curve with a triple point, when embedded in a noisy background ($90\%$ of background noise points). Left panel: noise points (dots)  (again for a run out of the $100$ we considered) and given curve (solid). Right panel: noise points (dots) and recognized  curves (solid); the only color present in the panel (magenta) means that  the repetition rate of the recognized curve is  higher than $50\%$.}}
\label{F3}
\end{center}
\end{figure}

In Figure \ref{F1} we represent the recognized curves in $100$ runs when the background noise is at $99\%$.
The colors of the curves are associated to their repetition rates, as follows. Cyan, from $2\%$ to $3\%$; green,  from $3\%$ to $5\%$; yellow,  from $5\%$ to $10\%$;  orange, from $10\%$ to $20\%$; red, from $20\%$ to $50\%$; magenta,  higher than $50\%$.
In Figure \ref{F2}  we show the recognized curves with  background noise  at $95\%$ level: almost all the recognitions are perfect with the exception of the curve shown in panel (c) where in $3\%$ of the cases at most, a profile not perfectly matching the given curve is found. In Figure \ref{F3} we  show the recognized quartic curve with a triple point  when the background noise is  decreased to  $90\%$ (the only case which  seemed critical at the previously considered noise level). 

 As previously stated, in all the trials the cyan and green curves occurred with a repetition rate lower than $5\%$ and  for this reason we can assume they are not stable, reliable  estimations of the real parameters.  Further, as experimentally shown in Tables \ref{T2} and \ref{T3},
the results look stable  for  background noise  starting from $90\%$, so we omit the tables and figures corresponding to the   $x=80,85$ cases.

\begin{table}[h]
\begin{center}
\begin{tabular}{|c||c|c|c|c|}
\hline 
\multirow{2}{*}{{\em Family of curves}}
 &  \multicolumn{3}{|c|}{{\em Background noise level}}  \\
\cline{2-4}
 & \multicolumn{1}{|c|}{$99\%$}& \multicolumn{1}{|c|}{$95\%$} & \multicolumn{1}{|c|}{$90\%$} \\
 \hline
 \hline 
 Descartes Folium & $41\%$ &$100\%$ &  $100\%$ \\
 \hline 
 Elliptic curve & $100\%$ & $100\%$ &  $100\%$  \\
 \hline 
 Quartic curve with triple point & $14\%$& $98\%$ & $100\%$\\
  \hline
 Quartic curve with tacnode &$92\%$& $100\%$ &  $100\%$ \\
  \hline 
 \end{tabular}
\end{center}
\caption{\small{Percentage of runs (out of 100) in which we exactly recognize the parameters for the four curves and for different  levels of background noise.}}
\label{T2}
\end{table}

\begin{table}[h]
\begin{center}
\begin{tabular}{|c||c|c|c|c|}
\hline 
\multirow{2}{*}{{\em Family of curves}}
 &  \multicolumn{3}{|c|}{{\em Background noise level}}  \\
\cline{2-4}
 & \multicolumn{1}{|c|}{$99\%$}& \multicolumn{1}{|c|}{$95\%$} & \multicolumn{1}{|c|}{$90\%$} \\
 \hline
 \hline 
 Descartes Folium   & $0.6 \pm 0.6$ & $0 \pm 0$  & $0\pm0$ \\
 \hline 
 Elliptic curve & $0 \pm 0$ & $0 \pm 0$  & $0\pm0$ \\
 \hline 
 Quartic curve with triple point & $0.4 \pm 0.4$ & $0.0002\pm0.0014$  & $0\pm0$ \\
  \hline
 Quartic curve with tacnode & $0.2 \pm 1.0$ & $0\pm0$ & $0\pm0$\\
  \hline 
 \end{tabular}
\end{center}
\caption{\small{Average distances, and corresponding  standard  deviations, between the pair of exact parameters and the recognized ones for the  four curves and for different  levels of background noise.}}
\label{T3}
\end{table}

\subsubsection{The case of the  Descartes Folium}\label{CaseDF}
Tables \ref{T2} and \ref{T3} show that the recognition of the Descartes Folium  is not  completely reliable in the case of $99\%$ of background noise points. In \cite{BMP}, the recognition of the Descartes Folium with this background noise percentage  was performed  by using a total of $10000$ points $N_2 = 9900$ and $N_1 = 100$. Here we  want to investigate how much we need to increase the $N_1$ value (from the initial $\nu_{{\rm opt}}$ value), and consequently the total number of points $N$, in order to have perfectly reliable recognitions even with $99\%$ of background noise points. We employ the same procedure as in the  previous section. In Table \ref{TFolium} we show the number of runs (out of $100$) in which the method correctly recognizes the parameters of the Descartes Folium for different values of  $N_1$, and then $N$.  As we can see it is necessary to increase the $N_1$ value to $25$ in order to have a $100\%$ of correct recognitions in presence of $99\%$ of background noise points.

\begin{table}[h]
\begin{center}
\begin{tabular}{|c|c|c|c|}
\hline
$N_1\ge\nu_{{\rm opt}}$ & $N_2$ & $N=N_1+N_2$ & {\em Percentage of runs}\\
\hline \hline
$9 $ & $891$ & $ 900 $ & $41\%$\\
\hline
$15 $ & $1485$ & $ 1500$ & $92\%$\\
\hline
$20 $ & $1980$ & $ 2000$ & $97\%$\\
\hline
$25 $ & $2475$ & $ 2500$ & $100\%$\\
\hline
\end{tabular}
\end{center}
\caption{\small{Percentage of runs (out of 100) in which we exactly recognize the parameters for the Descartes Folium with $99\%$ of background noise points  by increasing the number of points selected on the given curve with respect to the bound value $\nu_{{\rm opt}}=9$}.}
\label{TFolium}
\end{table}

\subsection{Robustness against random perturbation of points' locations}\label{RPP}

Here we validate the robustness of the recognition method against random perturbations of the location of points on the curves, following the procedure already employed in  \cite[p. 405]{BMP},  and by using $\nu_{\rm opt}$ instead of $N=100$. The procedure is repeated for $100$ runs.
More specifically, for each of the four families of curves as above:
\begin{enumerate}
\item Take $\nu_{\rm opt}$ points randomly on the curve, according to a uniform distribution. 
\item Repeat for $100$ different runs the steps:
\begin{enumerate}
\item perturb each coordinate of each point $(x,y)$ in this database by means of a Gaussian distribution $\mathcal{N}(0,\sigma^2)$ with zero mean and standard deviation $\sigma$;
\item  apply the recognition algorithm and determine the pair of parameters characterizing the curve;
\item In the parameter space, compute the Euclidean distance between the computed parameter pair and the exact one.
\end{enumerate}
\item Compute the average value, and corresponding standard deviations, of the $100$ distances computed in step $2$(c).
\item Repeat the procedure from step $2$ for a different value of the standard deviation in the Gaussian distribution.
\end{enumerate}

The results of this test are shown in Table \ref{T4}. First,  note that the recognition capability of the method in the case of random perturbations  of the  points' locations on the curve  deteriorates differently for the four curves: the elliptic curve and the quartic curve with tacnode show poor results starting from $\sigma=0.04$, while  in the case of the other curves  the algorithm performs relatively well even with $\sigma = 0.15$. 
Next, we also look at the number of exact recognitions of the four given curves.  The Descartes Folium behaves well for small values of $\sigma$, with  exact recognition rates of   $24\%$,   $12\%$, $4\%$, $1\%$, $1\%$ for $\sigma = 0.01,~0.02,~0.04,~0.05,~0.06$, respectively.  These values may seem low but, if combined with those shown in Table~\ref{T4}, they indicate that even when not perfect the recognition is still very accurate. The elliptic curve  case shows high rates of exact recognition ($89\%$, $44\%$, $14\%$, $10\%$, $4\%$, $1\%$ for $\sigma = 0.01,~0.02,~0.04,~0.05,~0.06,~0.1$, respectively), but, at the same time,  when the recognition goes wrong, the parameters values we found rather differ  from $a=-4$, $b=7$, also in the case of small values of the standard deviation $\sigma$,  thus justifying the overall non-optimal behavior shown in Table~\ref{T4}. 
In the case of the  quartic curve with triple point, we never find  the exact parameters $a=\frac{1}{5}$, $b=\frac{1}{2}$,  but we get parameters values  rather close to them for all the considered values of $\sigma$.
The  quartic curve with tacnode presents  an $11\%$ rate of exact recognition for  $\sigma = 0.01$, while for higher values of $\sigma$ the recognition of the exact parameters $a = 1$, $b = 8$ systematically  fails.

\begin{table}[h]
\begin{center}
\begin{tabular}{c|c|c|c|c|}
\cline{2-5}
&  \multicolumn{4}{|c|}{ {\em Family of curves}}  \\
\hline
\multicolumn{1}{|c||}{\em Standard} & \multicolumn{1}{|c|}{Descartes} & 
 \multicolumn{1}{|c|}{Elliptic} &
 \multicolumn{1}{|c|}{Quartic curve} &
 \multicolumn{1}{|c|}{ Quartic curve} \\
\multicolumn{1}{|c||}{\em deviation $\sigma$} & \multicolumn{1}{|c|}{Folium} & 
 \multicolumn{1}{|c|}{curve} &
 \multicolumn{1}{|c|}{with triple point } &
 \multicolumn{1}{|c|}{with tacnode } \\
 \hline
\hline
\multicolumn{1}{|c||}{$0.01$} & $0.08 \pm 0.1$ & $0.2\pm0.7$ & $0.02\pm0.03$ & $0.1\pm0.1$\\
\multicolumn{1}{|c||}{$0.02$} & $0.2 \pm 0.6$ & $0.6\pm1.6$ & $0.06\pm0.09$  & $0.4\pm0.8$\\
\multicolumn{1}{|c||}{$0.04$} & $0.4\pm 0.7 $ &  $1.1\pm1.9$ & $0.4\pm0.6$ & $0.8\pm1.2$\\
\multicolumn{1}{|c||}{$0.05$} & $0.4 \pm 0.6$ & $1.5\pm2.4$ & $0.5\pm0.6$ & $1.0\pm1.3$\\
\multicolumn{1}{|c||}{$0.06$} & $0.3 \pm 0.5$ & $2.4\pm3.1$ & $0.5\pm0.6$ & $1.3\pm1.7$ \\
\multicolumn{1}{|c||}{$0.08$} & $0.5\pm 0.6$ & $2.6\pm3.2$ & $0.6\pm0.8$ & $1.8\pm1.8$ \\
\multicolumn{1}{|c||}{$0.1$} & $0.4\pm 0.4$ & $3.4\pm3.4$ & $0.7\pm0.7$ & $2.2\pm2.2$\\
\multicolumn{1}{|c||}{$0.15$} & $0.4\pm 0.4$ & $4.0\pm3.2$ & $0.7\pm0.6$  & $2.9\pm$2.1\\
\hline
\end{tabular}
\end{center}
\caption{\small{Results of a test assessing  the robustness of the recognition method with respect to random perturbations of  points' locations on the four curves: the rows contain the average distance, and corresponding standard deviation, between  the pair of the exact parameters and the pair of the recognized  ones.}}
\label{T4}
\end{table}

\begin{figure}[h]
\begin{center}
\begin{tabular}{ccc}
\includegraphics[height=4cm]{fig_ris_descartes_true_nopt} &\includegraphics[height=4cm]{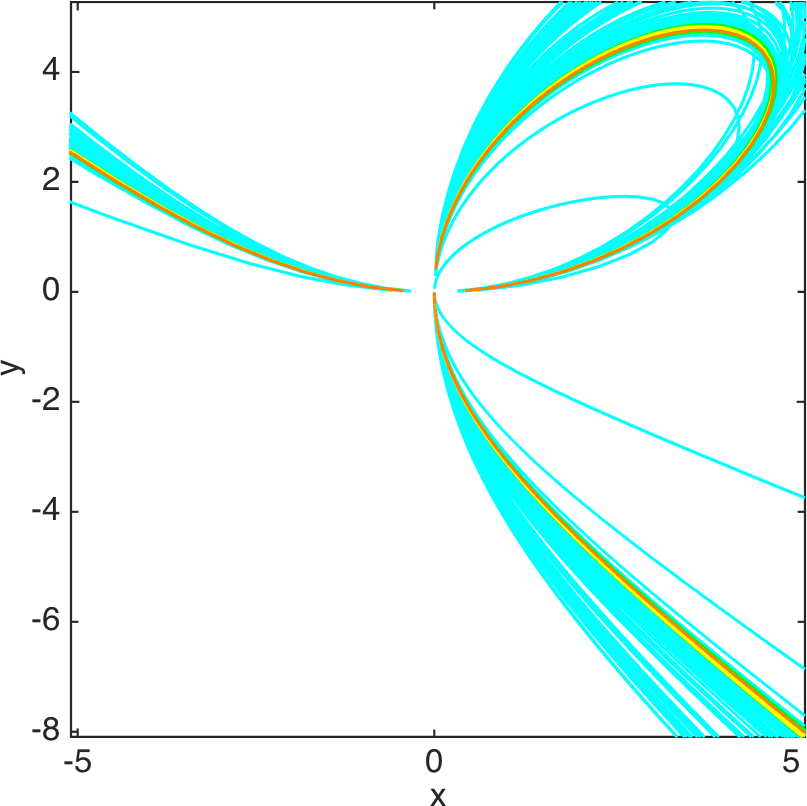} & \includegraphics[height=4cm]{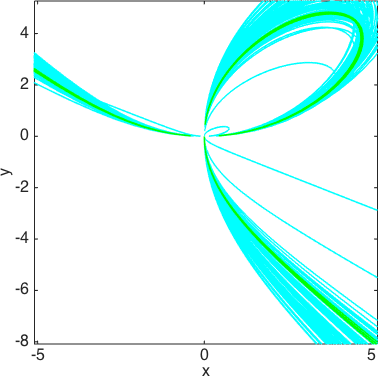}  \\
\includegraphics[height=4cm]{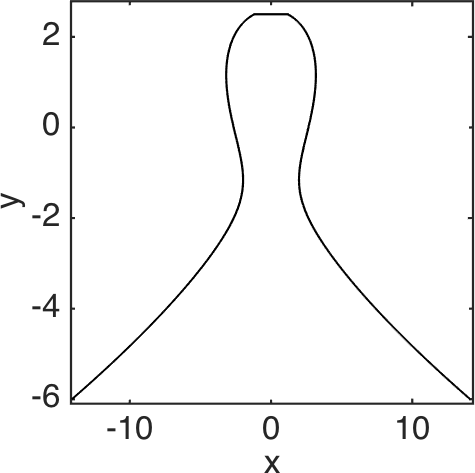} &\includegraphics[height=4cm]{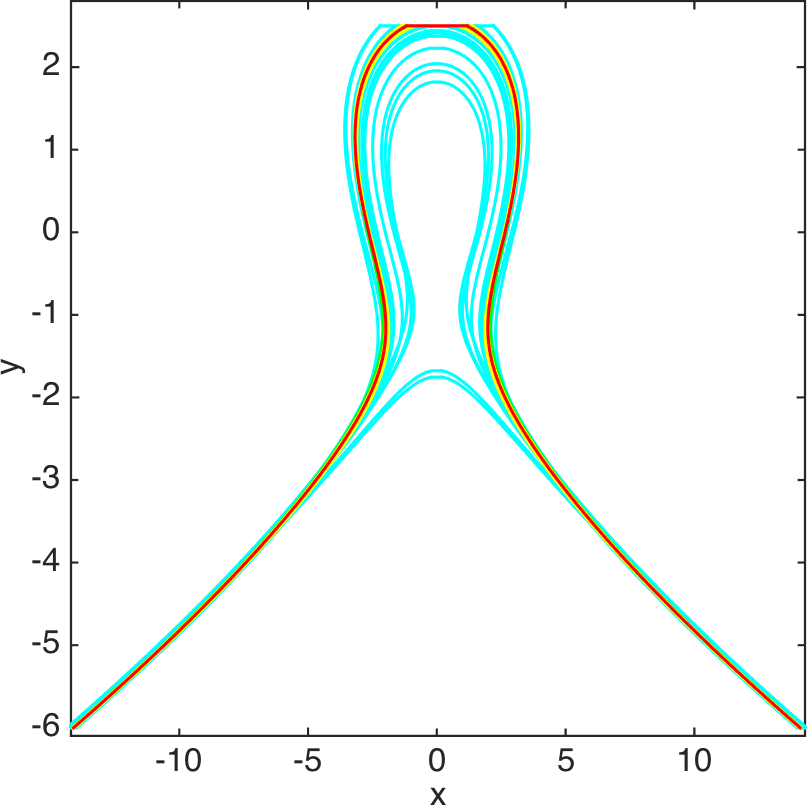} & \includegraphics[height=4cm]{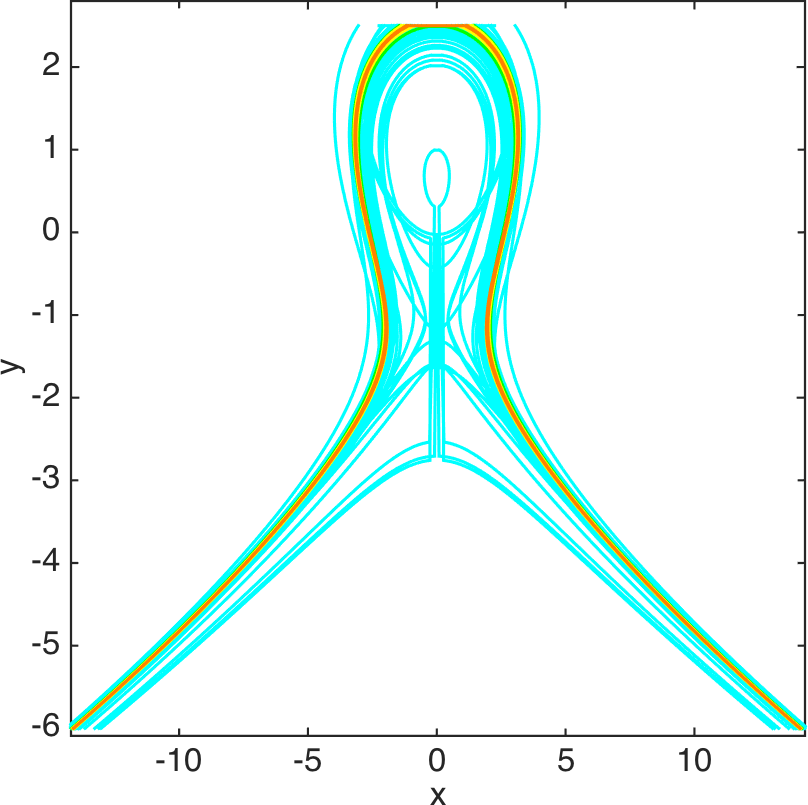}\\
\includegraphics[height=4cm]{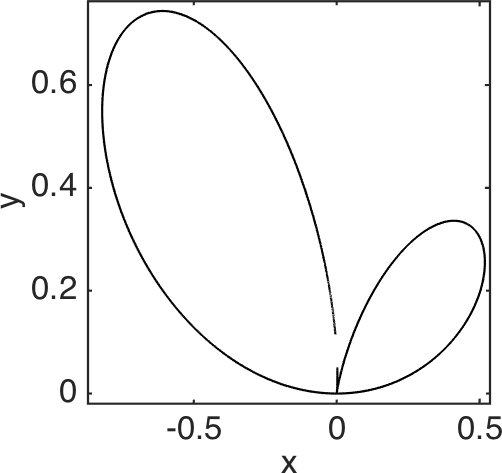} &\includegraphics[height=4cm]{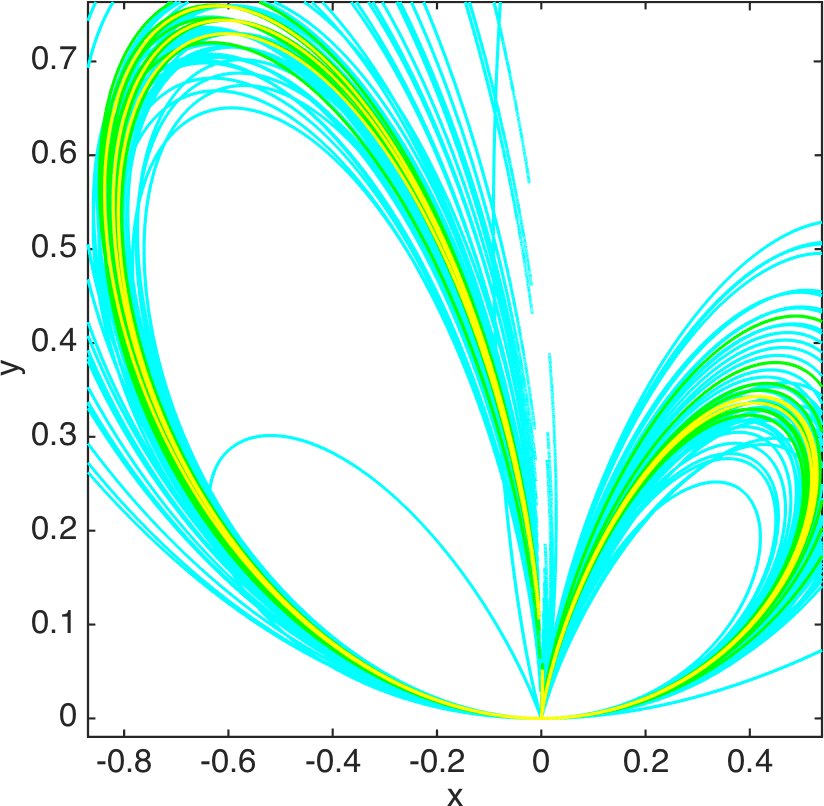} & \includegraphics[height=4cm]{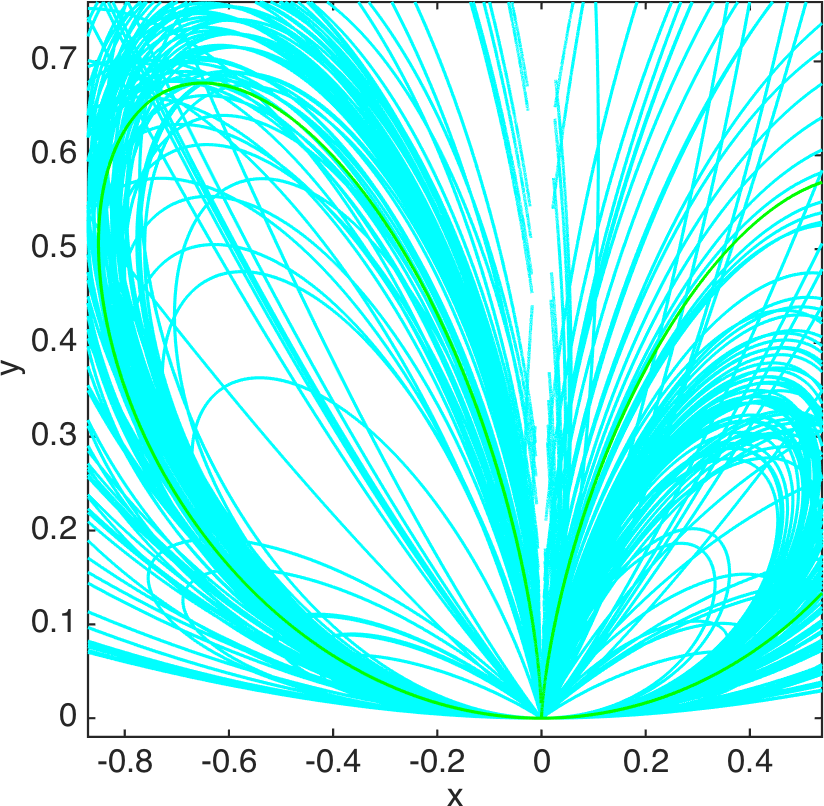}\\
\includegraphics[height=4cm]{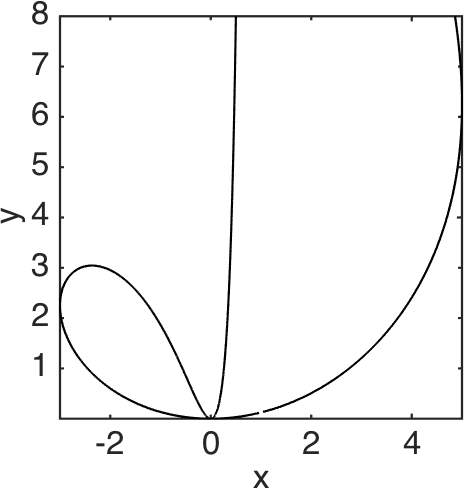} &\includegraphics[height=4cm]{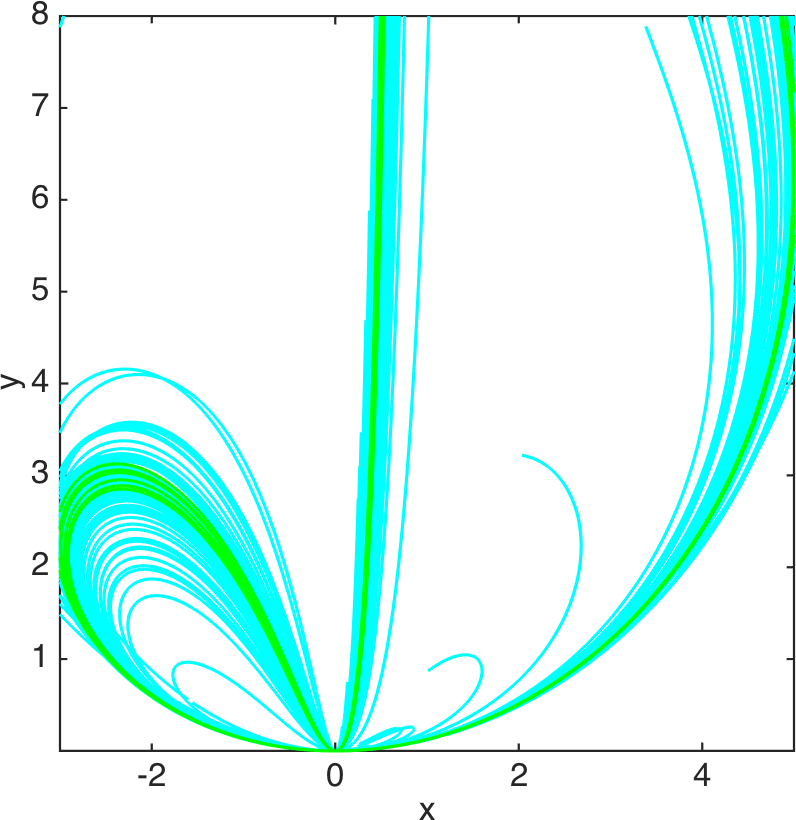} & \includegraphics[height=4cm]{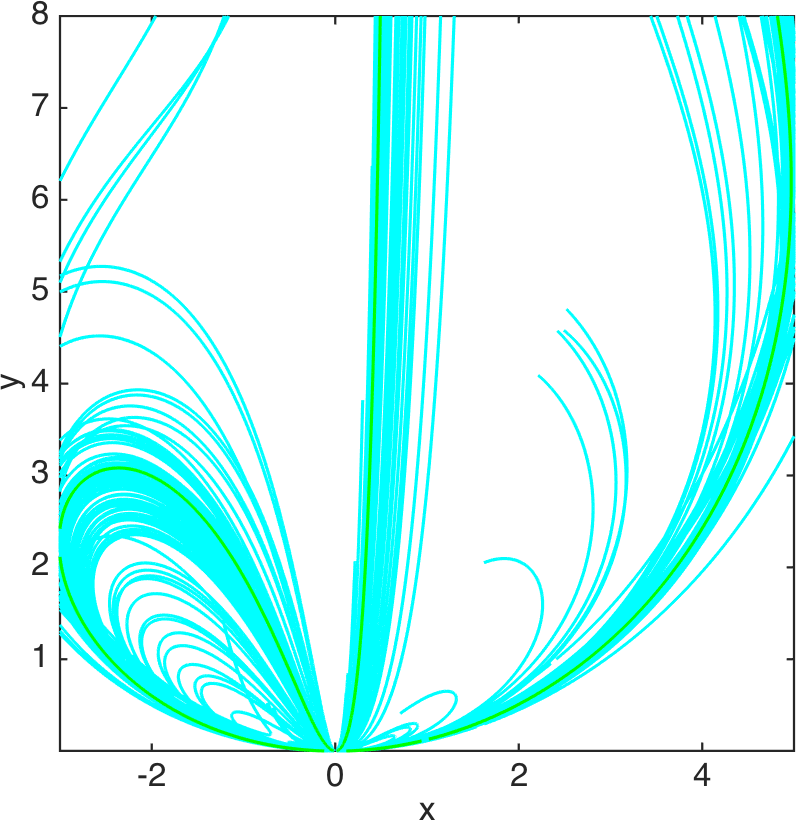}
\end{tabular}
\caption{\small{The  four given curves (left column) and their recognition when  the locations of $\nu_{\rm opt}$  points  randomly taken on the curves are perturbed  by using a $\mathcal{N}(0,\sigma^2)$ distribution where $\sigma = 0.02$ (central column) and $\sigma = 0.04$ (right column). The colors of the curves are associated to their repetition rates: from $2\%$ to $3\%$: cyan;  from $3\%$ to $5\%$: green;  from $5\%$ to $10\%$: yellow;  from $10\%$ to $20\%$: orange; from $20\%$ to $50\%$: red.}}
\label{F4}
\end{center}
\end{figure}

In Figure \ref{F4} we summarize the recognized curves in the $ 100$  runs when $\sigma = 0.02 $ (central column) and $\sigma = 0.04$ (right column). The colors of the curves are associated with the repetition rates as above. The repetition rates of the curves are rather low in most of the cases (the cyan color is associated with a repetition rate between the $2\%$ and the $3\%$), except for the elliptic curve, in which the given curve associated to $a=-4$, $b=7$ has a repetition rate from  $20\%$ to $50\%$ (red color) for $\sigma = 0.02$ and  a repetition rate from $10\%$ to $20\%$ (orange color) for $\sigma = 0.04$. We note that the  quartic curve with tacnode shows a  higher average distance between  the pair of the exact parameters $a=1$, $b=8$ and the pairs of the recognized parameters  if compared with the other curves; however,  the  graphs of the curves associated to the  recognized parameters look ``reasonably close" to that of the given quartic  curve (see Figure \ref{F4}, last row).

\subsection{Computational cost}\label{CC}

 The use of bound  $\nu_{\rm{opt}}$ allows us to recognize the curve by using a relatively small number of ``good" points. Here we assess the reduction of the computational cost when a small set of points is  considered: for each family of curves, we measured the time needed to go through the recognition procedure  in the case of $99\%$ background noise (the same percentage as in \cite{BMP}) with the same number of points as  in Section \ref{RNB} and in \cite{BMP}.  The results, provided in Table \ref{Ttime}, show a significant decrease in terms of time for all the families of curves we considered here  with a minimum factor $3.7$ and a maximum $21.2$.

\begin{table}[h]
\begin{center}
\begin{tabular}{|c||c|c||c|c|}
\hline 
{\em Family of curves} & $N$ & time [s] & $N$& time [s] \\
\hline\hline 
 Descartes Folium   & $900$ & $1.5$  & $10000$ & $12.7$\\
 \hline 
 Elliptic curve & $900$ & $5.1$  & $15800$ & $108.5$ \\
 \hline 
 Quartic curve with triple point & $1400$ & $3.5$  & $10000$ & $21.5$  \\
  \hline
 Quartic curve with tacnode & $1300$ & $7.9$ & $5000$ & $29.6$ \\
  \hline 
 \end{tabular}
\end{center}
\caption{\small{Comparison between computational times (in seconds) for the recognition algorithm, with background noise level at $99\%$, when the number of employed points is  $N=\nu_{\rm opt}+N_2$ (first two columns), and when $N$ is as in \cite[Section 6]{BMP}  (third and fourth columns).   }}
\label{Ttime}
\end{table}

\section{Conclusions}\label{Conc}
We propose a finite bound, $\nu_{\rm opt}$ (see Proposition \ref{A&Cprop}), for the number of  transforms to be considered in the accumulator function step of the recognition algorithm on which the Hough transform technique is based. Such a bound looks quite reliable and definitely of potential interest to reduce the computational burden associated to the accumulator function computation and optimization.  In particular, we obtain quite effective results  when the curves are embedded in a noisy background not exceeding  $95\%$ (figures \ref{F2} and \ref{F3}).  E.g., for background noise at $95\%$, with a data set of $N_1=\nu_{\rm opt}$   and $N_2=\frac{95}{5}\nu_{\rm opt}=  19\nu_{\rm opt}$  noise points (see relation (\ref{formula})), we recognize the curve by considering a total  of
$$N=N_1+N_2=20 \nu_{\rm opt}\approx 20d^2$$
Hough transforms, where $d$ denotes the degree of the curves from the family. Note that in \cite[Section 6, Table I]{BMP} the recognition is extremely effective even when  the curves are embedded in a  noisy background at  $99\%$, but using a number $N$ of total Hough transforms which approximately ranges from $5\times 10^3$ to $15\times 10^3$. 

Not surprisingly, the results are not  as  good  against  random  perturbations of points' locations on the curves. In the case of the quartic curve with a tacnode (the only one explicitly shown in \cite[Table 2]{BMP}),  we may  for instance note that  by using $N_1=\nu_{\rm opt}=13$ (instead of $N_1=100$) we  need a standard deviation $\sigma=0.01$ (instead of $0.04$) to get  the same average distance and corresponding standard deviation $0.1\pm 0.1$. Even if our results  get worse as $\sigma$ increases, they deserve to be noted.

\appendix
\section{An algebraic bound}\label{AB}
We keep the notation and assumptions  as in the previous sections.
A better understanding  of the behavior of equations  defining the  Hough transforms in the parameter space leads to   a refinement of Proposition \ref{A&Cprop} (see Proposition \ref{finally}).
 
 \medskip
 To begin with, let's add some comments on the degree and dimension of the Hough transform of points in ${\mathbb A}_{\x}^n(K)$, $K=\R,\C$.

%
Clearly, there exists a Zariski open set    ${\mathcal{U}}_1\subseteq {\mathbb A}_{\x}^n(K)$ such that, for each point $p\in {\mathcal{U}}_1$, the Hough transform $\Gamma_p({\mathcal{F}}): f_p(\bm \Lambda)=0$  of $p$   is a zero locus of a polynomial   of degree $h$ (not depending on $p$) in the parameter space. Since the Euclidean topology is finer than the Zariski topology, this holds true on a Euclidean open set ${\mathcal{U}}_1$ as well.
If $K=\C$, 
the Hough transform  $\Gamma_p({\mathcal{F}})$ is  a hypersurface. If $K=\R$, then  $\Gamma_p(\mathcal{F})$ is  $(t-1)$-dimensional  if and only if the  polynomial $f_p=f_p(\bm \Lambda)\in \R[\Lambda_1,\ldots ,\Lambda_t]$ has a non-singular zero in $\bm\lambda\in \R^t$, that is, the gradient
$\Big(\frac{\partial f_p}{\partial \Lambda_1}(\bm\lambda), \ldots,    \frac{\partial f_p}{\partial \Lambda_t}(\bm\lambda)   \Big)\neq 0$
(see again \cite[Theorem 4.5.1]{BCR} for details and equivalent conditions). A standard argument then shows that 
   there exists a Euclidean open set ${\mathcal{U}}_2\subseteq {\mathbb A}_{\x}^n(\R)$ such that for each point $p\in {\mathcal{U}}_2$ the Hough transform $\Gamma_p({\mathcal{F}})$ is a hypersurface in ${\mathbb A}_{\bm\Lambda}^t(\R)$ (for instance, see \cite{TBSe} for details). Indeed, as a special case of a more general result (see \cite[Proposition 2.25]{Ro}),  it holds true that  
  the Hough transform  $\Gamma_p(\mathcal{F})$  is $(t-1)$-dimensional  for a  generic point $p\in  {\mathbb A}_{(x,y)}^2(K)$, if $K$ is a field.
The above comments amount to conclude that,  for each point $p$ varying in the Euclidean open set ${\mathcal{U}}_1\cap{\mathcal{U}}_2\subseteq {\mathbb A}_{\x}^n(K)$,  the Hough transform $\Gamma_p({\mathcal F})$ is a hypersurface  of given degree $h$ not depending on $p$. Following \cite[Section 4]{TBSe}  we then define   the  {\em Hough transforms invariance degree open set} 
 as ${\mathcal{U}}_1$ if $K=\C$ and ${\mathcal{U}}_1\cap {\mathcal{U}}_2$ if $K=\R$. 
 
 \medskip
From now on, we assume $n=2$. First, we note  a  fact  we subsume  in the sequel. Let ${\mathcal B}_{\rm aff}$  be the base locus associated  to a family ${\mathcal F}=\{{\mathcal C}_{\bm\lambda}\}$ of curves (see Definition \ref{defB}). Since clearly
 ${\mathcal{U}}_1\cap{\mathcal B}_{\rm aff}=\emptyset$,  one has
 $${\mathcal C}_{\bm\lambda}\cap{\mathcal{U}}_1\subseteq {\mathcal C}_{\bm\lambda}\setminus{\mathcal B}_{\rm aff}$$
 for each curve ${\mathcal C}_{\bm\lambda}$ from the family.

Given a
point $p=(x_p,y_p)$  in the image space, belonging  to the invariance degree open set ${\mathcal U}_1\subset{\mathbb A}_{(x,y)}^2(K)$, write the polynomial $f_p(\bm \Lambda)$,  defining the Hough transform $\Gamma_p(\mathcal F)$ of  $p$, as
\begin{equation}\label{HTP}
f_p(\bm\Lambda) = \sum_{i+j=0}^dx_p^iy_p^j~g_{ij}(\bm\Lambda)=\sum_{m_1,\ldots,m_t} f_{m_1,\ldots,m_t}(x_p,y_p) \Lambda_1^{m_1} \ldots \Lambda_t^{m_t} \in K[\bm\Lambda],
\end{equation}
with $0 \le m_1+\cdots+m_t \le h$, where $h$ is the degree of $f_p(\Lambda)$. 
Let $f_{p}(\bm \Lambda)=f_h+\cdots+f_0$ be the decomposition
of $f_{p}(\bm \Lambda)$ into homogeneous components, where  $f_\gra \in K[\bm \Lambda]$ 
is homogeneous of degree $\gra$, for $\gra=0,\ldots,h$.  
Let  $\Lambda_0$ be the new  homogenizing coordinate.
The homogenization of $f_{p}(\bm \Lambda)$ with respect to $\Lambda_0$ is the polynomial 
$f_{p}(\bm \Lambda)^{\hom}=f_h+f_{h-1}\Lambda_0+\cdots+f_0\Lambda_0^h 
\in K[\Lambda_0,\bm \Lambda]$. 

We order the monomials  of the polynomial ring $K[\Lambda_0,\Lambda_1,\ldots,\Lambda_t]$; for instance, according to the {\tt degree-lexicographic} order with $\Lambda_t<\cdots <\Lambda_0 $ (see \cite[p. 48]{CA}). Let's give some definitions.

\begin{definition*}\label{OS} We say that the set
$
{\mathscr S}=\bigcup_{p\in {\mathcal U}_1}\big({\rm Supp}(f_p(\bm\Lambda))^{\rm hom}\big)
$
is the {\em generic ordered support}   according to the fixed ordering.
We also write $s:=\#{\mathscr S}$. 
\end{definition*}

\begin{definition*} \label{Mmatrix} Take a finite set of points $p_1,\ldots,p_\nu$ in the image space, belonging  to the invariance degree open set ${\mathcal U}_1$, and let 
$
M(p_1,\ldots,p_\nu;{\mathcal F})\in  {\rm Mat}_{\nu\times s}(K)
$
 be the matrix whose $j$-th row consists of  the coefficients of the polynomial $f_{p_j}(\bm \Lambda)^{\hom}$
ordered according to Definition \ref{OS}. We  say that $M(p_1,\ldots,p_\nu;{\mathcal F})$
is the HT-{\em matrix associated to  the points $p_1,\ldots,p_\nu$ with respect to  the family ${\mathcal F}$}. We denote by $\varrho(M(p_1,\ldots,p_\nu;{\mathcal F}))$ its rank.
\end{definition*}

We  are interested to find  a minimal set of generators of the ideal $ \big(f_{p_1}(\bm \Lambda), \ldots,f_{p_\nu}(\bm \Lambda)\big)$ in $K(\bm\Lambda]$. To this purpose,  just  for technical reasons we pass  to  the homogenization, then working in  
$K[\Lambda_0,\bm \Lambda]$. The following general fact (not involving specific curves from the family) achieves our goal.

\begin{propdef}\label{laura2} 
  Notation as above.
Let ${\mathcal F}$ be a family of curves  in ${\mathbb A}_{(x,y)}^2(K)$. 
Let ${\mathscr I}=\{p_1,\ldots,p_\nu\}$ be a set of distinct points belonging  to the invariance degree open set ${\mathcal U}_1\subset{\mathbb A}_{(x,y)}^2(K)$. 
 Let ${\mathscr T}_\nu:=\bigcap_{i=1, \ldots, \nu} \Gamma_{p_j}(\mathcal F)$. Consider the smallest positive integer
$\nu_{\rm best}:=\nu_{\rm best}(p_1,\ldots,p_\nu)\leq \nu$ defined  by the condition that there exist indices
$1\leq j_1 \le \cdots \le j_{\nu_{\rm best}}\leq \nu$ such that
$$\big(f_{p_1}(\bm \Lambda)^{\hom},\ldots, f_{p_\nu}(\bm \Lambda)^{\hom}\big)=
\big(f_{p_{j_1}}(\bm \Lambda)^{\hom},\ldots, f_{p_{j_{\nu_{\rm best}}}}(\bm \Lambda)^{\hom}\big),$$
and  set ${\mathscr T}_{\rm best}:= \Gamma_{p_{j_1}}(\mathcal F)\cap\ldots \cap
\Gamma_{p_{j_{\nu_{\rm best} }}}(\mathcal F)$.
Then, 
\begin{enumerate}
\em\item\em  $\nu_{\rm best}=\varrho\big(M(p_1,\ldots,p_\nu;{\mathcal F})\big)$;
\em\item\em ${\mathscr T}_\nu={\mathscr T}_{\rm best}$.
\end{enumerate}
\end{propdef}

\proof To prove statement $1)$, set $M:=M(p_1,\ldots,p_\nu;{\mathcal F})$ and 
 let  us first show $\nu_{\rm best}\leq  \varrho(M)$.  If $\varrho(M)=\nu$, then obviously $\nu_{\rm best}\leq \varrho(M)$, so we assume that $\varrho(M)<\nu$.
 We know that there exist $\varrho(M)$ rows of $M$ which are linearly independent and span the vectors space generated by all the rows of $M$. Up to renaming, we can assume that these  are  the first $\varrho(M) $  rows of $M$. Pick the $j$-th row $R_j$ of $M$ with $j>\varrho(M)$. Then $R_j$ can  be written as a linear combination of the rows $R_1,\ldots,R_{\varrho(M) }$. That is (denoting  for simplicity  $f_{p_j}:=f_{p_j}(\bm\Lambda)$, $j=1,\ldots,\nu$), there exist $\gra_1^j,\ldots, \gra_{\varrho(M) }^j\in K$ such that
$R_j=\gra_1^jR_1+\cdots+\gra_{\varrho(M) }^jR_{\varrho(M)}$.
This implies that $f_{p_j}^{\hom}=\gra_1^jf_{p_1}^{\hom}+\cdots+\gra_{\varrho(M) }^jf_{p_{\varrho(M)}}^{\hom}$, so  that
$f_{p_j}^{\hom}\in \big(f_{p_1}^{\hom},\ldots,f_{p_{\varrho(M)}}^{\hom}\big)$.
Since this holds true for each $j>\varrho(M)$, we have
$$ \big(f_{p_1}^{\hom},\ldots,f_{p_{\nu}}^{\hom}\big)\subseteq\big(f_{p_1}^{\hom},\ldots,f_{p_{\varrho(M)}}^{\hom}\big),$$
which implies 
$ \big(f_{p_1}^{\hom},\ldots,f_{p_{\nu}}^{\hom}\big)=\big(f_{p_1}^{\hom},\ldots,f_{p_{\varrho(M)}}^{\hom}\big)$. We then conclude that $\nu_{\rm best}\leq  \varrho(M)$ by the minimality of $\nu_{\rm best}$.

To show the converse, and  up to renaming, let $f_{p_1}^{\hom},\ldots,f_{p_{\nu_{\rm best}}}^{\hom}$ be the generators of $ \big(f_{p_1}^{\hom},\ldots,f_{p_{\nu}}^{\hom}\big)$. If $\nu_{\rm best}=\nu$ there is nothing to prove, so we assume that $\nu_{\rm best}<\nu$.
 For each $f_{p_j}^{\hom}$ with $j>\nu_{\rm best}$ we have $f_{p_j}^{\hom}\in \big(f_{p_1}^{\hom},\ldots,f_{p_{\nu_{\rm best}}}^{\hom}   \big)$, that is, there exist polynomials $h_i^j$, $j=1,\ldots,\nu$, $i=1,\ldots,\nu_{\rm best}$, such that
\begin{equation}\label{L1}
f_{p_j}^{\hom}=h_1^jf_{p_1}^{\hom}+\cdots+h_{\nu_{\rm best}}^jf_{p_{\nu_{\rm best}}}^{\hom}. 
\end{equation}
Since $f_{p_j}^{\hom}$  and $f_{p_1}^{\hom},\ldots,f_{p_{\nu_{\rm best}}}^{\hom}$ are homogeneous polynomials of the same degree it follows that the $h_i^j$'s  are homogeneous  of degree zero, that is, $h_i^j\in K$. Thus, equality  (\ref{L1}) is equivalent to say  that each row $R_j$ of $M$ is a linear combination of $R_1,\ldots,R_{\nu_{\rm best}}$. The conclusion $\varrho(M)\leq \nu_{\rm best}$ then immediately follows. 

As to statement $2)$, 
consider the ideal $(f_{p_1}(\bm \Lambda),\ldots,  f_{p_\nu}(\bm \Lambda))$ in $K[\bm \Lambda]$. We want to prove that there exist indices $j_1,\ldots,j_{\rm \nu_{\rm best}}$, $1\leq j_1< \cdots <j_{\rm \nu_{\rm best}}\leq \nu$, such  that
$$(f_{p_1}(\bm \Lambda),\ldots,  f_{p_\nu}(\bm \Lambda))=\big(f_{p_{j_1}}(\bm \Lambda), \ldots,f_{p_{j_{\rm \nu_{\rm best}  }}}(\bm \Lambda)\big).$$
The inclusion ``$\supseteq$'' is obvious, so we only have to prove the converse inclusion ``$\subseteq$". By definition of $\nu_{\rm best}$, we know that there are indices $1\leq j_1< \cdots <j_{\rm \nu_{\rm best}}\leq \nu$, such  that
\begin{eqnarray*}
(f_{p_1}^{\rm hom}(\bm \Lambda),\ldots,  f_{p_\nu}^{\rm hom}(\bm \Lambda)) &=& 
\big(f_{p_{j_1}}^{\rm hom}(\bm\Lambda), \ldots,f_{p_{j_{\rm \nu_{\rm best}  }}}^{\rm hom}(\bm \Lambda)\big)  \\
&\subseteq&\big(f_{p_{j_1}}(\bm \Lambda), \ldots,f_{p_{j_{\rm \nu_{\rm best}  }}}(\bm \Lambda)\big)^{\rm hom},
\end{eqnarray*}
where the last inclusion follows  by definition of ideal homogenization (see \cite[Definition 4.3.4]{KR05}).
Passing to the dehomogenization, we get (see \cite[Proposition  4.3.12]{KR05})
\begin{eqnarray*}
(f_{p_1}^{\rm hom}(\bm \Lambda),\ldots,  f_{p_\nu}^{\rm hom}(\bm \Lambda))^{\rm deh} &= &
\big(f_{p_{j_1}}^{\rm hom}(\bm\Lambda), \ldots,f_{p_{j_{\rm \nu_{\rm best}  }}}^{\rm hom}(\bm \Lambda)\big)^{\rm deh}  \\
&\subseteq&\Big(\big(f_{p_{j_1}}(\bm \Lambda), \ldots,f_{p_{j_{\rm \nu_{\rm best}  }}}(\bm \Lambda)\big)^{\rm hom}\Big)^{\rm deh}\\
&=&
\big(f_{p_{j_1}}(\bm \Lambda), \ldots,f_{p_{j_{\rm \nu_{\rm best}  }}}(\bm \Lambda)\big),
\end{eqnarray*}
where the last equality is a consequence of \cite[Proposition 4.3.5]{KR05}. Since 
$$(f_{p_1}^{\rm hom}(\bm \Lambda),\ldots,  f_{p_\nu}^{\rm hom}(\bm \Lambda))^{\rm deh}
=
\big(f_{p_1}(\bm \Lambda), \ldots,f_{p_{\nu}}(\bm \Lambda)\big),$$ 
(see \cite[Corollary 4.3.8]{KR05}), the claimed inclusion follows. Thus, we can conclude that
$${\mathscr T}_\nu=\Gamma_{p_1}(\mathcal F)\cap\ldots \cap
\Gamma_{p_\nu}(\mathcal F)=
\Gamma_{p_{j_1}}(\mathcal F)\cap\ldots \cap
\Gamma_{p_{j_{\nu_{\rm best} }}}(\mathcal F).$$
\qed

\begin{prop}\label{finally}
  Notation as above.
  Let ${\mathcal F}=\{{\mathcal C}_{{\bm \lambda}}\}$ be a family of  curves  in ${\mathbb A}_{(x,y)}^2(K)$.  
  Fix a curve ${\mathcal C}_{{\bm \lambda}}$ from the family, and take   $ \nu_{\rm opt}=d^2-\#{\mathcal B}(\C)+1$ distinct points  $p_1,\ldots,p_{\nu_{\rm opt}}$  on $
  {\mathcal C}_{{\bm \lambda}}\cap {\mathcal U}_1$. 
  Let ${\mathscr T}=\cap_{p \in C_{\bm \lambda}} \Gamma_p(\mathcal{F})$ and  
 let ${\mathscr T}_{\rm best} = \bigcap_{j=1, \ldots,    \nu_{\rm best}}   \Gamma_{p_j}(\mathcal F)$.
 Then we have:
\begin{enumerate}
\em\item\em ${\mathcal C}_{{\bm \lambda}'}={\mathcal C}_{{\bm \lambda} }$ for each ${\bm \lambda}' \in {\mathscr T}_{\rm best}$.
\em\item\em ${\mathscr T}_{\rm best}={\mathscr T}$.
\em\item\em  If  the family ${\mathcal F}$  is Hough regular, then ${\mathscr T}_{\rm best}=\{{\bm \lambda}\}$.
\end{enumerate}
\end{prop}
\proof If $ \nu_{\rm best}=d^2-\#{\mathcal B}(\C)+1$, the result simply  follows from Proposition \ref{A&Cprop}. 
Then  we can assume that
$\nu_{\rm best}=\varrho(M(p_1,\dots,p_{\nu_{\rm opt}};{\mathcal F}))< d^2-\#{\mathcal B}(\C)+1$. Therefore,  Proposition-Definition \ref{laura2}(2)  yields
$$ {\mathscr T}_{\rm opt}= \bigcap_{j=1,\ldots,\nu_{\rm opt} } \Gamma_{p_j}(\mathcal F)= \bigcap_{j=1, \ldots,    \nu_{\rm best}  } \Gamma_{p_j}(\mathcal F)= {\mathscr T}_{\rm best}.$$
Thus, Proposition \ref{A&Cprop} applies again to  conclude the proof.
\qed

The following remark clarifies the relations between the bounds $\nu_{\rm opt}$ (see Proposition~\ref{A&Cprop}) and  $\nu_{\rm best}$, as 
well as,  for families $\mathcal F$ which are Hough regular,  between them and the number of parameters~$t$.

\begin{rem*}\label{relations} Assumptions and notation as in Proposition \ref{finally}. In fact,  instead of $\nu_{\rm best}$, it is possible to use the easier computable bound
$$
\nu_{\rm best}':= \min\{s-1,d^2-\#{\mathcal B}(\C)+1\},
$$
with $s=\#{\mathscr S} $ as in Definition \ref{OS}.
This follows from the fact that the points $p_j$, 
$j=1,\ldots,\nu_{\rm opt}$, lie on a given  curve ${\mathcal C}_{\bm\lambda}$, $\bm\lambda=(\lambda_1,\ldots,\lambda_t)$, from the family ${\mathcal F}$, and consequently the $s$  columns of $M$ are linearly dependent. Precisely, recalling expression (\ref{HTP}), one has
$$f_{p_j}(\bm\Lambda) = 
\sum_{m_1,\ldots,m_t} f_{m_1,\ldots,m_t}(x_{p_j},y_{p_j}) \Lambda_1^{m_1} \ldots \Lambda_t^{m_t} =0,$$
and the $j$-th row of $M$ is made up of the coefficients (ordered according to the fixed ordering) $ f_{m_1,\ldots,m_t}(x_{p_j},y_{p_j}) $, for $j=1,\ldots,\nu_{\rm opt}$. 
In conclusion, we have  the inequalities:
\begin{equation}\label{nuova} \nu_{\rm best}=\varrho(M)\leq   \nu_{\rm best}'= \min\{s-1,d^2-\#{\mathcal B}(\C)+1\}\leq \nu_{\rm opt}=d^2-\#{\mathcal B}(\C)+1.\end{equation}

Now,  assume that the family $\mathcal F$ is Hough regular. Coming back to Subsection \ref{PS}, consider the ideal
$$I=\big(f_{p_1}(\bm \Lambda), \ldots,f_{p_h}(\bm \Lambda)\big)
\subset K[\Lambda]$$
generated by the polynomials $f_p(\bm \Lambda)$  defining the Hough transforms $\Gamma_p({\mathcal F})$,
 $p\in {\mathcal C}_\lambda$.
Let $m$ be the minimal number of generators of $I$ in $K[\Lambda]$, so that, as we noted, $h\geq m$. 
Furthermore, such a number $m$ has to satisfy the lower bound
$m \ge \textrm{codim}(I)$ (see  \cite{HR} and also \cite[Chapter 10]{CA}). By definition, 
$\textrm{codim}(I) := \textrm{dim}(K[\Lambda]) - \textrm{dim}(K[\Lambda]/I)$. Since $\textrm{dim}(K[\Lambda])=t$
and $\textrm{dim}(K[\Lambda]/I)=0$ (this derives from the assumption that the family $\mathcal F$ is Hough regular,
which implies that the ideal $I$ is zero-dimensional), it then follows $m\geq t$, whence $h\geq t$. Thus,  in particular, relations (\ref{nuova}) yield
$$\nu_{\rm opt} \geq \nu_{\rm best}\geq t.$$
\quadrato\end{rem*}

We provide here some illustrative examples in the real case.

\begin{example*} (Curve of Lamet)\label{lamet}
Consider in $\mathbb A^2_{(x,y)}(\mathbb R)$ the  family ${\mathcal F}=\{{\mathcal C}_{a,b}\}$  of curves  of degree $m$  of equation
$\left(\frac{x}{a}\right)^m+\frac{y^m}{b}=1$, that is,
\begin{equation}\label{Q7}
 \mathcal{C}_{a,b}: bx^m+a^my^m=a^mb,\end{equation}
  for positive real numbers $a$, $b$. The {\em curve of Lamet}  is clearly non-singular (even in the complex projective plane ${\mathbb P}^2(\comp)$), and then of  genus $\frac{(m-1)(m-2)}{2}$.
  
We further assume that  the degree $m$ is  even. As noted  below, this assures  the boundedness  of the curve. (If $m$ is odd  the curve is unbounded: think, for example, to the Fermat elliptic cubic of equation $x^3+y^3=1$.)
 Indeed, the knowledge of some basic facts about $p$-norms on $\mathbb R^n$
 allows us to show that  the curve of Lamet is contained in the rectangular  region
\begin{eqnarray*}
\left \{(x,y) \in \mathbb A^2_{(x,y)}(\mathbb R) 
\: \big|\: -a \le x \le a, \: -b^{1/m} \le y \le b^{1/m} \right \}.
\end{eqnarray*}

Passing to homogeneous coordinates we have
$$\overline{{\mathcal C}_{a,b}}:  bx_0^m+a^mx_1^m-a^mbx_2^m=0,  $$ whence ${\mathcal B}(\C)=\{\emptyset\}$.
In order to compute $\nu_{\rm best}$, note that
for any point $p=(x_p,y_p)$  in the invariance degree open set ${\mathcal U}_1\subset {\mathbb A}_{(x,y)}^2(\reals)$, the Hough transform is the $(m+1)$-degree curve $\Gamma_p({\mathcal F})$ in the parameter plane $\langle A, B \rangle$ of equation
$$
f_p(A,B)= A^mB -y_p^mA^m-x_p^mB=0.$$
Therefore the  bound from  Proposition \ref{A&Cprop}   becomes
$$\nu_{\rm opt}=d^2-\#{\mathcal B}(\C)+1=m^2+1.$$

For instance, in the case $m=4$, $\nu_{\rm opt}=17$. 
Thus, Proposition  \ref{finally} yields
 $$ \nu_{\rm best}\leq\min\{s-1,d^2-\#{\mathcal B}(\C)+1\} =\min\{2,17\}=2,$$ where $s:=\#{\mathscr S}$. To see that $\nu_{\rm best}=2$, take $17$ points $p_\ell=(x_{p_\ell}, y_{p_\ell})$, $\ell=1,\ldots,17$,  on the Lamet curve
$$ \mathcal{C}_{a,b}: bx^4+a^4y^4-a^4b=0,$$  with $a$, $b$ \underline{fixed}  and the points belonging to the open set 
${\mathcal U}_1$.

The coefficient of the maximum degree term of  $f_p(A,B)$ equals $1$, so that it generates the whole ring $\R[A,B]$, that is, ${\mathcal U}_1=\R^2$.  Keeping the notation as in the proof of  Proposition  \ref{finally},
 consider the (transpose of)  HT-matrix $M$  associated to the set of points $\{p_\ell\}_{\ell=1,\ldots, 17}$, that is, 
$$
M^t =\left( \begin{matrix} 1 & 1 & 1 & \cdots  & 1 &  \cdots  &           1                      \\
{\Area -y_{p_1}^4} & -y_{p_2}^4& -y_{p_3}^4 &  \cdots &  -y_{p_\ell}^4 &  \cdots  &    -y_{p_{17}}^4       \\
{\Area -x_{p_1}^4} & -x_{p_2}^4& -x_{p_3}^4 & \cdots &  -x_{p_\ell}^4  &  \cdots  &  -x_{p_{17}}^4       \\
\end{matrix} \right) \in {\rm Mat}_{3\times 17}(\R).$$

Compute, for $i, j,k\in \{1,\ldots,17\}$, $i\neq j\neq k$,

\begin{eqnarray*}
\left | \begin{matrix} 1 & 1 & 1    \\
{\Area -y_{p_i}^4} & -y_{p_j}^4& -y_{p_k}^4 \\
{\Area -x_{p_i}^4} & -x_{p_j}^4& -x_{p_k}^4   &    \\
\end{matrix} \right | &=& 
\left | \begin{matrix} 1 & 0 & 0    \\
{\Area -y_{p_i}^4} & y_{p_i}^4-y_{p_j}^4 &y_{p_i}^4 -y_{p_k}^4 \\ 
{\Area -x_{p_i}^4}  & x_{p_i}^4-x_{p_j}^4 &x_{p_i}^4 -x_{p_k}^4  &     \\
\end{matrix} \right | \\
&=& (x_{p_i}^4-x_{p_k}^4)(y_{p_i}^4-y_{p_j}^4)  -(x_{p_i}^4-x_{p_j}^4)(y_{p_i}^4-y_{p_k}^4)\\
&=&\frac{1}{b}(y_{p_i}^4-y_{p_j}^4)(bx_{p_i}^4-bx_{p_k}^4)-\frac{1}{b}(y_{p_i}^4-y_{p_k}^4)(bx_{p_i}^4-bx_{p_j}^4)\\
&=&\frac{1}{b}\big[(y_{p_i}^4-y_{p_j}^4)(-a^4y_{p_i}^4+a^4y_{p_k}^4)
-(y_{p_i}^4-y_{p_k}^4)(-a^4y_{p_i}^4+a^4y_{p_j}^4)\big] \\
&=&\frac{a^4}{b}\big[(y_{p_i}^4-y_{p_j}^4)(-y_{p_i}^4+y_{p_k}^4)
-(y_{p_i}^4-y_{p_k}^4)(-y_{p_i}^4+y_{p_j}^4)\big]=0,
\end{eqnarray*}
to conclude that  $\varrho(M)=\nu_{\rm best}=2$. \quadrato
\end{example*}

\begin{example*}\label{Ex0}
Consider in $\mathbb A^2_{(x,y)}(\mathbb R)$  the family ${\mathcal F}=\{{\mathcal C}_{a,b}\}$ of conics of equation
$$
{\mathcal C}_{a,b}:a^2x^2+by+x=0,
$$ 
for  real parameters ${\bm \lambda}=(a,b)$. Passing to homogeneous coordinates  we see that $\#{\mathcal B}(\C)=2$. Whence $\nu_{\rm opt}=d^2-\#{\mathcal B}(\C)+1=3$. For a general point $p=(x_p,y_p)$,  the Hough transform is the conic of equation 
$$f_p(A,B)=x_p^2A ^2+y_pB+x_p=0,$$ 
so that  $s=\#{\rm Supp}(f_p(A,B))=3=\nu_{\rm opt}$.
Consider the three points 
$p_1=(1,-2)$, $p_2=(-1,0)$, $p_3=(-2,-2)$ on ${\mathcal C}_{1,1}$
and the polynomials
$f_{p_1}(A,B)=A^2-2B+1$, $f_{p_2}(A,B)=A^2-1$, $f_{p_3}(A,B)=4A^2-2B-2$.
 The HT-matrix $M \in \rm{Mat}_{3 \times 3}(\R)$ is
$$
M=\left( \begin{matrix}  1 & -2 & 1  \\1 & 0 & -1 \\ 4 & -2 & -2
\end{matrix} \right),
$$ 
 whose rank is $\nu_{\rm{best}}=\varrho(M)=2$.  We have ${\mathscr T}_{\rm best}=\Gamma_{p_1}({\mathcal F})\cap\Gamma_{p_2}({\mathcal F})=\{(1,1), (-1,1)\}$. (Note that $\Gamma_{p_1}({\mathcal F})$, $\Gamma_{p_2}({\mathcal F})$ have two  more  coinciding  common points  at infinity).
The family ${\mathcal F}$ is not Hough regular unless $a>0$, in which case  ${\mathscr T}_{\rm best}=\{(1,1)\}$.
\quadrato

\end{example*}

Although most of the times $\varrho(M)<d^2-\#{\mathcal B}(\C)+1$, so that $\nu_{\rm best}=\rho(M)$
there are also cases  where the equality
$\varrho(M)=d^2-\#{\mathcal B}(\C)+1$ holds true, as the following simple example shows.

\begin{example*}\label{Ex1}
Consider  in $\mathbb A^2_{(x,y)}(\mathbb R)$  the family ${\mathcal F}=\{{\mathcal C}_{a,b,c}\}$ of lines of equation
$$
{\mathcal C}_{a,b,c}:ax+by+c=0,
$$ 
for  real parameters ${\bm \lambda}=(a,b,c)$. The polynomial defining the Hough transform of a general point $(x,y)$ is
$$
f_{(x,y)}(A,B,C) = xA+yB+C \in \R[x,y][A,B,C],
$$
having support ${\mathscr S}=\{A,B,C\}$. Then $s=3$, $d^2-\#{\mathcal B}(\C)+1=s-1=2$,
whence $\nu_{\rm best}\leq 2$. 
 Consider the two points 
$p_1=(0,-1)$, $p_2=(-1,0)$ on ${\mathcal C}_{1,1,1}$ , and the polynomials
$f_{p_1}(A,B,C)=-B+C$, $f_{p_2}(A,B,C)=-A+C$. The HT-matrix $M \in \rm{Mat}_{2 \times 3}(\R)$ is
$$
M = \left( \begin{matrix}
0 & -1 & 1\\ -1 & 0 & 1
\end{matrix}\right), $$
whose rank is $\nu_{\rm best }=\varrho(M)=2=\nu_{\rm opt}$.
Then the set ${\mathscr T}_{\rm best}=\Gamma_{p_1}({\mathcal F})\cap\Gamma_{p_2}({\mathcal F})$  coincides with the line $\{(t,t,t)\:|\:t \in \R\}$ 
in the parameter space $\R^3=\langle A, B, C\rangle$. Clearly, ${\mathcal C}_{t,t,t}={\mathcal C}_{1,1,1}$ for each $t \in \R$, according to Proposition \ref{finally}(1).

The family ${\mathcal F}$ is not Hough regular, meeting the regularity property as soon as one of the parameters is fixed. For instance, letting $c=1$, we get the family of lines ${\mathcal F}'=\{{\mathcal C}_{a,b}: ax+by+1=0\}$,  and now ${\mathscr T}_{\rm best}=\Gamma_{p_1}({\mathcal F}')\cap\Gamma_{p_2}({\mathcal F}')=\{(1,1)\}$ with $p_1=(0,-1)$, $p_2=(-1,0)$ on ${\mathcal C}_{1,1}$.

Since the parameters are linear, the same conclusions follow from Lemma  \ref{linlem}.
 \quadrato
\end{example*}

\bigskip
\noindent{\bf Acknowledgments} We would  like to thank our friend and   former colleague Annalisa Perasso, who previously effectively worked on  the first stages of the project.

\smallskip

\small{

}

\end{document}